\documentclass[letterpaper, 10 pt, conference]{ieeeconf}
\IEEEoverridecommandlockouts      
\usepackage{subfiles}
\usepackage{graphics} 
\usepackage{epsfig}   
\usepackage{caption} 
\usepackage{mathptmx} 
\usepackage{textcomp}
\usepackage{times}

\usepackage{amsmath}  
\usepackage{amsfonts}
\usepackage{amssymb}  
\usepackage{bbm}
\usepackage{xcolor}
\usepackage{graphicx}  
\usepackage{tabularx}
\usepackage{amsthm}
\usepackage{setspace}
\usepackage{soul}
\usepackage[ruled,vlined]{algorithm2e}
\DeclareMathOperator*{\argmin}{arg\,min\;}
\DeclareMathOperator*{\argmax}{arg\,max\;}
\DeclareMathOperator*{\suchthat}{\textrm{s.t.} \;}
\newcommand{\costtogo}[0]{\textit{cost\_to\_go}}
\newcommand{\constrainttogo}[0]{\textit{constraint\_to\_go}}
\newcommand{\local}{local }
\newcommand{\Local}{Local }
\usepackage{gensymb}
\usepackage{url}
\usepackage{multirow}
\usepackage{mathtools}
\let\labelindent\relax
\usepackage{enumitem}
\newlist{steps}{enumerate}{1}
\setlist[steps, 1]{label = \colorcounter{*}}
\usepackage{tikz}
\usetikzlibrary{decorations.text}
\usetikzlibrary{shapes}
\usetikzlibrary{arrows.meta}
\usetikzlibrary{decorations.markings}
\usepackage{adjustbox}
\usepackage{cite} 
\usepackage{hyperref}
\usepackage{calc} 

\makeatletter
\let\MYcaption\@makecaption
\makeatother

\usepackage[font=footnotesize]{subcaption}

\makeatletter
\let\@makecaption\MYcaption
\makeatother

\newlength\replength
\newcommand\repfrac{.1}

\newlength\dashpaddinglength
\newcommand\rulewidth{.1pt}
\newcommand\tdashfill[1][\repfrac]{\cleaders\hbox to \replength{%
  \smash{\rule[\arraystretch\ht\strutbox]{\repfrac\replength}{\rulewidth}}}\hfill}
\newcommand\tabdashline{%
  \makebox[\dashpaddinglength][r]{\makebox[\tabcolsep]{\tdashfill\hfil}}\tdashfill\hfil%
  \makebox[\dashpaddinglength][l]{\makebox[\tabcolsep]{\tdashfill\hfil}}%
  \\[-\arraystretch\dimexpr\ht\strutbox+\dp\strutbox\relax]%
}

\title{\LARGE \bf
Equality Constrained Linear Optimal Control With Factor Graphs
}

\definecolor{darkgreen}{rgb}{0, 0.5, 0}
\newcommand{\red}[1]{\textcolor{red}{#1}}
\newcommand{\green}[1]{\textcolor{darkgreen}{#1}}
\newcommand{\blue}[1]{\textcolor{blue}{#1}}
\makeatletter
\newcommand{\colorcounter}[1]{\expandafter\@colorcounter\csname c@#1\endcsname}
\newcommand{\@colorcounter}[1]{%
    \ifcase#1
    \or\red{1.}
    \or\green{2.}
    \or\blue{3.}
    \else error! \fi%
}
\AddEnumerateCounter{\colorcounter}{\@colorcounter}{m}
\makeatother
\newif\ifcrossnew

\author{Shuo Yang\textsuperscript{1},
        Gerry Chen\textsuperscript{2},
        Yetong Zhang\textsuperscript{2},
        Howie Choset\textsuperscript{1},
        and Frank Dellaert\textsuperscript{2}
        
\thanks{\textsuperscript{1} Shuo Yang and Howie Choset are with the Robotics Institute and Department of Mechanical Engineering, Carnegie Mellon University, Pittsburgh. Emails: \texttt{\{shuoyang, choset\}@andrew.cmu.edu} }

\thanks{\textsuperscript{2} Gerry Chen,
        Yetong Zhang,
        and Frank Dellaert are with the Institute for Robotics and Intelligent Machines, Georgia Institute of Technology, Atlanta. Emails: \texttt{\{gchen328, yzhang3333, fd27\}@gatech.edu} }
        
}

\begin{document}

\maketitle
\begin{abstract}
This paper presents a novel factor graph-based approach to solve the discrete-time finite-horizon Linear Quadratic Regulator problem subject to auxiliary linear equality constraints within and across time steps. 
We represent such optimal control problems using constrained factor graphs and optimize the factor graphs to obtain the optimal trajectory and the feedback control policies using the variable elimination algorithm with a modified Gram-Schmidt process. We prove that our approach has the same order of computational complexity as the state-of-the-art dynamic programming approach. 
Furthermore, current dynamic programming approaches can only handle equality constraints between variables at the same time step, but ours can handle equality constraints among any combination of variables at any time step while maintaining linear complexity with respect to trajectory length. Our approach can be used to efficiently generate trajectories and feedback control policies to achieve periodic motion or repetitive manipulation. 
\end{abstract}


\section{Introduction}\label{sec:intro}
The Equality Constrained Linear Quadratic Regulator (EC-LQR) is an important extension \cite{sideris2011riccati, laine2019efficient} of the Linear Quadratic Regulator (LQR) \cite{kalman1960new}.
The standard finite-horizon discrete-time LQR problem contains (1) quadratic costs on the state trajectory and the control input trajectory and (2) \textit{system dynamics constraints}
which enforce that the current state is determined by a linear function of the previous state and control.  In the EC-LQR, \textit{auxiliary constraints} are introduced to enforce additional linear equality relationships on one or more state(s) and/or control(s).  


In many important problems auxiliary constraints violate the Markov assumption, yet such constraints are rarely considered in existing EC-LQR approaches.
We classify auxiliary constraints in EC-LQR problems into two categories which we term \textit{\local constraints} and \textit{cross-time-step constraints}. A \local constraint only contains a state and/or control from the same time step. Examples of \local constraints include initial and terminal conditions on states, contact constraints, and states along a predefined curve. In contrast, a cross-time-step constraint involves multiple states and controls at different time instances.
Such non-Markovian constraints are pervasive in many robotics applications. For example, a legged robot's leg configuration must return to the same state after a period of time during a periodic gait \cite{farshidian2017real}. In optimal allocation with resource constraints \cite{boyd2004convex}, the sum of control inputs is constrained to be some constant. Our goal is to solve for both optimal trajectories and optimal feedback control policies in EC-LQR problems with \local and cross-time-step constraints in linear time with respect to the trajectory length, which no existing EC-LQR methods can achieve. 



Reformulating control problems as inference problems \cite{levine2018reinforcement, toussaint2009robot, watson2020stochastic, kappen2009optimal} is a growing alternative to common trajectory optimization \cite{kelly2017introduction,dai2014whole,posa2016optimization} and dynamic programming (DP) approaches for optimal control \cite{li2004iterative,sideris2011riccati,laine2019efficient} .  While trajectory optimization focuses on open-loop trajectories rather than feedback laws, and a method using DP to handle cross-time-step constraints has yet to be proposed, control as inference may offer the advantages of both.  Factor graphs, in particular, are a common tool for solving inference problems \cite{dellaert2017factor} and have recently been applied to optimal control \cite{dong2016motion, gerryyetong2019lqrfg}.

\begin{figure}
    \centering
    \resizebox{\linewidth}{!}{\begin{tikzpicture}
\coordinate (x0_coord) at (0.0, 0);
\coordinate (x1_coord) at (1.2, 0);
\coordinate (x2_coord) at (2.4, 0);
\coordinate (u0_coord) at (0.6, -1);
\coordinate (u1_coord) at (1.7999999999999998, -1);
\coordinate (soft_x0_coord) at (0.48, 0.2);
\coordinate (soft_x1_coord) at (1.68, 0.2);
\coordinate (soft_x2_coord) at (2.88, 0.2);
\coordinate (soft_u0_coord) at (0.12, -1.2);
\coordinate (soft_u1_coord) at (1.32, -1.2);
\coordinate (dynamics01_coord) at (0.6, -0.3333333333333333);
\coordinate (dynamics12_coord) at (1.8, -0.3333333333333333);
\coordinate (constrain0_coord) at (0.3, -0.5);
\coordinate (constrain1_coord) at (1.5, -0.5);
\coordinate (constrain2_coord) at (2.6999999999999997, -0.5);
\coordinate (cross_constraint_coord) at (1.2, 0.7);
\path[draw=black, line width=0.3pt] (x0_coord) -- (soft_x0_coord);
\path[draw=black, line width=0.3pt] (x1_coord) -- (soft_x1_coord);
\path[draw=black, line width=0.3pt] (x2_coord) -- (soft_x2_coord);
\path[draw=black, line width=0.3pt] (u0_coord) -- (soft_u0_coord);
\path[draw=black, line width=0.3pt] (u1_coord) -- (soft_u1_coord);
\path[draw=black, line width=0.3pt] (x0_coord) -- (dynamics01_coord);
\path[draw=black, line width=0.3pt] (u0_coord) -- (dynamics01_coord);
\path[draw=black, line width=0.3pt] (x1_coord) -- (dynamics01_coord);
\path[draw=black, line width=0.3pt] (x1_coord) -- (dynamics12_coord);
\path[draw=black, line width=0.3pt] (u1_coord) -- (dynamics12_coord);
\path[draw=black, line width=0.3pt] (x2_coord) -- (dynamics12_coord);
\path[draw=black, line width=0.3pt] (x0_coord) -- (constrain0_coord);
\path[draw=black, line width=0.3pt] (u0_coord) -- (constrain0_coord);
\path[draw=black, line width=0.3pt] (x1_coord) -- (constrain1_coord);
\path[draw=black, line width=0.3pt] (u1_coord) -- (constrain1_coord);
\path[draw=black, line width=0.3pt] (x2_coord) -- (constrain2_coord);
\path[draw=red, line width=0.3pt] (x0_coord) -- (cross_constraint_coord);
\path[draw=red, line width=0.3pt] (x2_coord) -- (cross_constraint_coord);
\node[scale=0.5, fill=white][circle, inner sep=2.8pt, draw, very thin] at (x0_coord) (x0) {$x_0$};
\node[scale=0.5, fill=white][circle, inner sep=2.8pt, draw, very thin] at (x1_coord) (x1) {$x_1$};
\node[scale=0.5, fill=white][circle, inner sep=2.8pt, draw, very thin] at (x2_coord) (x2) {$x_2$};
\node[scale=0.5, fill=white][circle, inner sep=2.8pt, draw, very thin] at (u0_coord) (u0) {$u_0$};
\node[scale=0.5, fill=white][circle, inner sep=2.8pt, draw, very thin] at (u1_coord) (u1) {$u_1$};
\node[circle, scale=0.3, fill=black] at (soft_x0_coord) (soft_x0) {};
\node[circle, scale=0.3, fill=black] at (soft_x1_coord) (soft_x1) {};
\node[circle, scale=0.3, fill=black] at (soft_x2_coord) (soft_x2) {};
\node[circle, scale=0.3, fill=black] at (soft_u0_coord) (soft_u0) {};
\node[circle, scale=0.3, fill=black] at (soft_u1_coord) (soft_u1) {};
\node[rectangle, scale=0.3, fill=black] at (dynamics01_coord) (dynamics01) {};
\node[rectangle, scale=0.3, fill=black] at (dynamics12_coord) (dynamics12) {};
\node[rectangle, scale=0.3, fill=black] at (constrain0_coord) (constrain0) {};
\node[rectangle, scale=0.3, fill=black] at (constrain1_coord) (constrain1) {};
\node[rectangle, scale=0.3, fill=black] at (constrain2_coord) (constrain2) {};
\node[rectangle, scale=0.3, fill=red] at (cross_constraint_coord) (cross_constraint) {};
\node[rectangle callout, draw=none, inner sep=2.8pt, rounded corners=1pt, fill = gray!20, callout absolute pointer={(soft_x0_coord)}, scale = 0.3] at (0.24,0.4) {$x_0^TQ_{xx_0}x_0$};
\node[rectangle callout, draw=none, inner sep=2.8pt, rounded corners=1pt, fill = gray!20, callout absolute pointer={(soft_x1_coord)}, scale = 0.3] at (1.44,0.4) {$x_1^TQ_{xx_1}x_1$};
\node[rectangle callout, draw=none, inner sep=2.8pt, rounded corners=1pt, fill = gray!20, callout absolute pointer={(soft_x2_coord)}, scale = 0.3] at (2.64,0.4) {$x_2^TQ_{xx_2}x_2$};
\node[rectangle callout, draw=none, inner sep=2.8pt, rounded corners=1pt, fill = gray!20, callout absolute pointer={(soft_u0_coord)}, scale = 0.3] at (0.12,-1.5) {$u_0^TQ_{uu_0}u_0$};
\node[rectangle callout, draw=none, inner sep=2.8pt, rounded corners=1pt, fill = gray!20, callout absolute pointer={(soft_u1_coord)}, scale = 0.3] at (1.32,-1.5) {$u_1^TQ_{uu_1}u_1$};
\node[rectangle callout, draw=none, inner sep=2.8pt, rounded corners=1pt, fill = gray!20, callout absolute pointer={(dynamics01_coord)}, scale = 0.3] at (0.6,0) {$x_1=F_{x_0}x_0+F_{u_0}u_0$};
\node[rectangle callout, draw=none, inner sep=2.8pt, rounded corners=1pt, fill = gray!20, callout absolute pointer={(dynamics12_coord)}, scale = 0.3] at (1.7999999999999998,0) {$x_2=F_{x_1}x_1+F_{u_1}u_1$};
\node[rectangle callout, draw=none, inner sep=2.8pt, rounded corners=1pt, fill = gray!20, callout absolute pointer={(constrain0_coord)}, scale = 0.3] at (-0.12,-0.7) {$G_{x_0}x_0+G_{u_0}u_0+g_{l_0}=0$};
\node[rectangle callout, draw=none, inner sep=2.8pt, rounded corners=1pt, fill = gray!20, callout absolute pointer={(constrain1_coord)}, scale = 0.3] at (1.08,-0.7) {$G_{x_1}x_1+G_{u_1}u_1+g_{l_1}=0$};
\node[rectangle callout, draw=none, inner sep=2.8pt, rounded corners=1pt, fill = gray!20, callout absolute pointer={(constrain2_coord)}, scale = 0.3] at (2.6999999999999997,-0.7) {$G_{x_2}x_2+g_{l_2}=0$};
\node[rectangle callout, draw=none, inner sep=2.8pt, rounded corners=1pt, fill = gray!20, callout absolute pointer={(cross_constraint_coord)}, scale = 0.3] at (1.7999999999999998,0.7) {$S_0x_0+S_2x_2+s=0$};
\coordinate (legend) at (2.4, -1.5);
\coordinate (legend1) at (2.5, -1.28);
\coordinate (legend2) at (2.5, -1.42);
\node[circle, scale=0.3, fill=black] at (legend1) {};
\node[draw=none, scale=0.3, anchor=west, inner sep=6pt] at (legend1) {Quadratic Objective Factor};
\node[rectangle, scale=0.3, fill=black] at (legend2) {};
\node[draw=none, scale=0.3, anchor=west, inner sep=6pt] at (legend2) {Linear Constraint Factor};
\end{tikzpicture}}
    \caption{\footnotesize{The factor graph representation of an Equality Constrained Linear Quadratic Regular (EC-LQR) problem. Circles with letters are states or controls. Filled squares and circles represent objectives and constraints that involve the state or controls to which they are connected. The red square represents a cross-time-step constraint.}}
    \label{fig:cross-ec-LQR-factor-graph}
\end{figure}
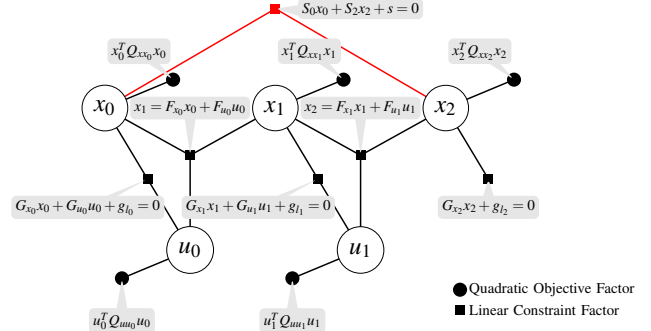

In this paper we propose a novel formulation using factor graphs \cite{dellaert2017factor} to efficiently solve the EC-LQR problem with both \local and cross-time-step constraints in linear time with respect to trajectory length. We demonstrate how to represent the EC-LQR problem as a factor graph (shown in Fig. \ref{fig:cross-ec-LQR-factor-graph}),
and apply the variable elimination (VE) algorithm \cite{blair1993introduction} on the factor graph to solve for the optimal trajectories and optimal feedback control policies.
The flexibility of the factor graph representation allows 
cross-time-step constraints with arbitrary numbers of variables to be seamlessly handled.
As long as the maximum time index difference of variables involved in each constraint is bounded, the computational complexity stays linear with trajectory length.
The approach in this paper matches the computational complexity of standard dynamic programming techniques [2], but also has the added benefit of handling cross-time-step constraints.


\section{Related Work}\label{sec:related}

Trajectory optimization methods typically transcript a problem into a Quadratic Programming (QP) \cite{barclay1998sqp} or NonLinear Programming (NLP) \cite{kelly2017introduction} problem which can be efficiently solved to obtain open-loop trajectories of nonlinear systems.  
Local controllers can be used to track 
the open-loop trajectories generated \cite{posa2016optimization}. Designing local controllers that obey equality constraints motivates EC-LQR problems.

For EC-LQR problems with just \local constraints, DP-based approaches can generate both the optimal trajectories and feedback control policies. Solving standard LQR using DP is well understood in control theory \cite{kalman1960new}. \cite{posa2016optimization} tackles EC-LQR with state-only \local constraints by projecting system dynamics onto the constraint manifold. \cite{sideris2011riccati} extends the DP approach by using Karush-Kuhn-Tucker (KKT) conditions \cite{boyd2004convex} to absorb auxiliary constraints into the cost function, but its computation time grows with the cube of the trajectory length for certain auxiliary constraints. \cite{laine2019efficient} solves the EC-LQR with \textit{\local}constraints in linear complexity by adding a new auxiliary constraint dubbed ``\constrainttogo{}'' at each time step during DP steps. 


Control as inference, in which a control problem is reformulated and solved as an inference problem, has gained considerable attention \cite{levine2018reinforcement, toussaint2009robot}.  Probabilistic Graphical Models (PGMs), which are commonly used for inference, have been applied to optimal control problems \cite{kappen2009optimal, watson2020stochastic} because they describe dependencies among variables while maintaining sparsity in the graphical representation. Therefore, PGMs can solve for variable distributions efficiently by exploiting sparsity \cite{koller2009probabilistic}.  The Markov assumption gives optimal control problems a ``chain'' structure when represented as PGMs allowing linear computational complexity with respect to trajectory length \cite{astrom1971introduction, toussaint2009robot, watson2020stochastic, gerryyetong2019lqrfg}, but PGMs can also exploit sparsity for more complex (non-chain) structures which motivates using PGMs for cross-time-step constraints.  

Factor graphs, a type of PGM, have been successfully applied to robot perception and state estimation \cite{dellaert2017factor}. Prior works have demonstrated that the variable elimination (VE) algorithm \cite{blair1993introduction} on factor graphs can efficiently factorize the graphs' equivalent matrix representations in order to infer the posterior distributions of random variables. This procedure is called \textit{factor graph optimization}. Moreover, factor graphs can encode constraints \cite{cunningham2010ddf}. Other than estimation, factor graphs can be used to do motion planning \cite{dong2016motion, ta2014factor}. Standard LQRs with factor graphs are considered in \cite{watson2020stochastic, gerryyetong2019lqrfg} without auxiliary constraints.

\section{Problem And Method}\label{sec:method}
In this section we first formulate the standard LQR and EC-LQR problems following the notation used in \cite{laine2019efficient}. 
Then we solve a standard LQR problem as a factor graph and review relevant concepts related to factor graphs. Next, we solve EC-LQR with \local constraints using factor graphs and compare our algorithm to the one proposed by \cite{laine2019efficient}, the most recent DP-based approach.
Finally, we show how our method handles EC-LQR with cross-time-step constraints.

\subsection{Problem Formulation}
For a robotic system with state $x_t \in \mathbb{R}^n$ and control input $u_t \in \mathbb{R}^m$, we define a state trajectory as $\textbf{x} = [x_0, x_1,\dots,x_T]$ and control input trajectory as $\textbf{u} = [u_0, u_1,\dots,u_{T-1}]$ where $T$ is the trajectory length. The optimal control input trajectory $\textbf{u}^*$ and its corresponding state trajectory $\textbf{x}^*$ are the solution to the constrained linear least squares problem:
\begin{subequations}\label{problem1}
\begin{align}
    \min_{\textbf{u}} & \ x_T^TQ_{xx_T}x_T + \sum_{t=0}^{T-1}(x_t^TQ_{xx_t}x_t+u_t^TQ_{uu_t}u_t) \label{eqn:linear_cost}\\
    \text{s.t. }      & \ x_{t+1} = F_{x_t}x_t + F_{u_t}u_t \label{eqn:linear_sys_dyn}\\
                      & \ G_{x_t}x_{t} + G_{u_t}u_{t} + g_{l_t} = 0, \ \ t \in \mathcal{C} \label{eqn:local_cst} \\
                      & \ G_{x_T}x_{T} + g_{l_T} = 0 \label{eqn:local_cst_final}\\
                      & \ \sum_{i\in C_{kx}} S_{xki}x_{i} + \sum_{j\in C_{ku}} S_{ukj}u_j + s_k = 0 \label{eqn:cross_cst} 
\end{align}
\end{subequations}
where
    $Q_{xx_T}$, $Q_{xx_t}$, and $Q_{uu_t}$ are positive definite matrices defining the cost function;
    $F_{x_t}$ and $F_{u_t}$ define the system dynamics at time $t$; 
    constraints (\ref{eqn:local_cst}) and (\ref{eqn:local_cst_final}) are \local auxiliary constraints; and constraint (\ref{eqn:cross_cst}) is a new formulation for cross-time-step constraints.
In (\ref{eqn:local_cst}) and (\ref{eqn:local_cst_final}),
    $G_{x_t} \in \mathbb{R}^{l_t\times n}$, $G_{u_t} \in \mathbb{R}^{l_t\times m}$, and $g_{l_t}\in \mathbb{R}^{l_t}$ form \local constraints with constraint dimension $l_t$;
    $\mathcal{C}$ is the set of time steps where a \local constraint, such as initial state constraint, applies; 
    and $G_{x_T}$ and $g_{l_T}$ form a \local constraint with dimension $l_T$ on the final step.
In the cross-time-step constraint (\ref{eqn:cross_cst}),
    %
    $S_{xki} \in \mathbb{R}^{c_t\times n}$,
    $S_{ukj} \in \mathbb{R}^{c_t\times m}$, and
    $s_k \in \mathbb{R}^{c_t}$ form constraints on a set of states $x_i$ and controls $u_j$ where $k$ is the index of the cross-time-step constraint. 
    In this paper we focus on representing quadratic cost in the factor graph, but linear terms in the cost function can be incorporated too as shown in the next section.
 

\subsection{Standard LQR as a Factor Graph} \label{sec:lqr}

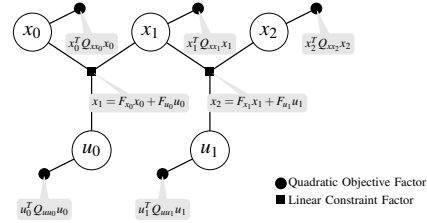
\begin{figure}
    \vspace{0.3em}
    \centering
        \centering
        \resizebox{0.67\linewidth}{!}{\begin{tikzpicture}
\coordinate (x0) at (0, 0);
\coordinate (x1) at (1, 0);
\coordinate (x2) at (2, 0);
\coordinate (u0) at (0.5, -1);
\coordinate (u1) at (1.5, -1);
\coordinate (f1) at (0.4, 0.2);
\coordinate (f2) at (1.4, 0.2);
\coordinate (f3) at (2.4, 0.2);
\coordinate (f4) at (0.1, -1.2);
\coordinate (f5) at (1.1, -1.2);
\coordinate (f6) at (0.5, -0.3333333333333333);
\coordinate (f7) at (1.5, -0.3333333333333333);
\path[draw, line width=0.3pt] (x0) -- (f1);
\path[draw, line width=0.3pt] (x1) -- (f2);
\path[draw, line width=0.3pt] (x2) -- (f3);
\path[draw, line width=0.3pt] (u0) -- (f4);
\path[draw, line width=0.3pt] (u1) -- (f5);
\path[draw, line width=0.3pt] (x0) -- (f6);
\path[draw, line width=0.3pt] (u0) -- (f6);
\path[draw, line width=0.3pt] (x1) -- (f6);
\path[draw, line width=0.3pt] (x1) -- (f7);
\path[draw, line width=0.3pt] (u1) -- (f7);
\path[draw, line width=0.3pt] (x2) -- (f7);
\node[scale=0.5, fill=white][circle, inner sep=2.8pt, draw, very thin] at (x0) {$x_0$};
\node[scale=0.5, fill=white][circle, inner sep=2.8pt, draw, very thin] at (x1) {$x_1$};
\node[scale=0.5, fill=white][circle, inner sep=2.8pt, draw, very thin] at (x2) {$x_2$};
\node[scale=0.5, fill=white][circle, inner sep=2.8pt, draw, very thin] at (u0) {$u_0$};
\node[scale=0.5, fill=white][circle, inner sep=2.8pt, draw, very thin] at (u1) {$u_1$};
\node[circle, scale=0.3, fill=black] at (f1) {};
\node[circle, scale=0.3, fill=black] at (f2) {};
\node[circle, scale=0.3, fill=black] at (f3) {};
\node[circle, scale=0.3, fill=black] at (f4) {};
\node[circle, scale=0.3, fill=black] at (f5) {};
\node[rectangle, scale=0.3, fill=black] at (f6) {};
\node[rectangle, scale=0.3, fill=black] at (f7) {};
\node[rectangle callout, draw=none, inner sep=2.8pt, rounded corners=1pt, fill = gray!20, callout absolute pointer={(f1)}, scale = 0.3] at (0.5,-0.1) {$x_0^TQ_{xx_0}x_0$};
\node[rectangle callout, draw=none, inner sep=2.8pt, rounded corners=1pt, fill = gray!20, callout absolute pointer={(f2)}, scale = 0.3] at (1.5,-0.1) {$x_1^TQ_{xx_1}x_1$};
\node[rectangle callout, draw=none, inner sep=2.8pt, rounded corners=1pt, fill = gray!20, callout absolute pointer={(f3)}, scale = 0.3] at (2.5,-0.1) {$x_2^TQ_{xx_2}x_2$};
\node[rectangle callout, draw=none, inner sep=2.8pt, rounded corners=1pt, fill = gray!20, callout absolute pointer={(f4)}, scale = 0.3] at (0.1,-1.5) {$u_0^TQ_{uu_0}u_0$};
\node[rectangle callout, draw=none, inner sep=2.8pt, rounded corners=1pt, fill = gray!20, callout absolute pointer={(f5)}, scale = 0.3] at (1.1,-1.5) {$u_1^TQ_{uu_1}u_1$};
\node[rectangle callout, draw=none, inner sep=2.8pt, rounded corners=1pt, fill = gray!20, callout absolute pointer={(f6)}, scale = 0.3] at (0.9,-0.6) {$x_1=F_{x_0}x_0+F_{u_0}u_0$};
\node[rectangle callout, draw=none, inner sep=2.8pt, rounded corners=1pt, fill = gray!20, callout absolute pointer={(f7)}, scale = 0.3] at (1.9,-0.6) {$x_2=F_{x_1}x_1+F_{u_1}u_1$};
\coordinate (legend) at (2, -1.5);
\coordinate (legend1) at (2.1, -1.28);
\coordinate (legend2) at (2.1, -1.42);
\node[circle, scale=0.3, fill=black] at (legend1) {};
\node[draw=none, scale=0.3, anchor=west, inner sep=6pt] at (legend1) {Quadratic Objective Factor};
\node[rectangle, scale=0.3, fill=black] at (legend2) {};
\node[draw=none, scale=0.3, anchor=west, inner sep=6pt] at (legend2) {Linear Constraint Factor};
\end{tikzpicture}}
    \caption{\footnotesize{Factor graph of a standard LQR problem with trajectory length $T=2$.}}
        \label{fig:lqr_suba}
\end{figure}

\begin{figure*}
    \setlength\tabcolsep{1pt}
    \centering
    \begin{subfigure}{0.53\linewidth}
    \vspace{0.4em}
        \centering
        \begin{tabular}{ccc}
             \adjustbox{valign=m}{\resizebox{1.1\width}{!}{\begin{tikzpicture}
\definecolor {dgreen} {rgb} { 0,0.5,0 };
\coordinate (x0_coord) at (0.0, 0);
\coordinate (x1_coord) at (1.2, 0);
\coordinate (x2_coord) at (2.4, 0);
\coordinate (u0_coord) at (0.6, -1);
\coordinate (u1_coord) at (1.7999999999999998, -1);
\coordinate (soft_x0_coord) at (0.48, 0.2);
\coordinate (soft_x1_coord) at (1.68, 0.2);
\coordinate (soft_x2_coord) at (2.88, 0.2);
\coordinate (soft_u0_coord) at (0.12, -1.2);
\coordinate (soft_u1_coord) at (1.32, -1.2);
\coordinate (dynamics01_coord) at (0.6, -0.3333333333333333);
\coordinate (dynamics12_coord) at (1.8, -0.3333333333333333);
\path[draw=black, line width=0.3pt] (x0_coord) -- (soft_x0_coord);
\path[draw=black, line width=0.3pt] (x1_coord) -- (soft_x1_coord);
\path[draw=red, line width=0.3pt] (x2_coord) -- (soft_x2_coord);
\path[draw=black, line width=0.3pt] (u0_coord) -- (soft_u0_coord);
\path[draw=black, line width=0.3pt] (u1_coord) -- (soft_u1_coord);
\path[draw=black, line width=0.3pt] (x0_coord) -- (dynamics01_coord);
\path[draw=black, line width=0.3pt] (u0_coord) -- (dynamics01_coord);
\path[draw=black, line width=0.3pt] (x1_coord) -- (dynamics01_coord);
\path[draw=red, line width=0.3pt] (x1_coord) -- (dynamics12_coord);
\path[draw=red, line width=0.3pt] (u1_coord) -- (dynamics12_coord);
\path[draw=red, line width=0.3pt] (x2_coord) -- (dynamics12_coord);
\node[scale=0.7, fill=white][circle, inner sep=2.8pt, draw, very thin] at (x0_coord) (x0) {$x_0$};
\node[scale=0.7, fill=white][circle, inner sep=2.8pt, draw, very thin] at (x1_coord) (x1) {$x_1$};
\node[scale=0.7, fill=white][circle, inner sep=2.8pt, draw, very thin] at (x2_coord) (x2) {$x_2$};
\node[scale=0.7, fill=white][circle, inner sep=2.8pt, draw, very thin] at (u0_coord) (u0) {$u_0$};
\node[scale=0.7, fill=white][circle, inner sep=2.8pt, draw, very thin] at (u1_coord) (u1) {$u_1$};
\node[circle, scale=0.5, fill=black] at (soft_x0_coord) (soft_x0) {};
\node[circle, scale=0.5, fill=black] at (soft_x1_coord) (soft_x1) {};
\node[circle, scale=0.5, fill=red] at (soft_x2_coord) (soft_x2) {};
\node[circle, scale=0.5, fill=black] at (soft_u0_coord) (soft_u0) {};
\node[circle, scale=0.5, fill=black] at (soft_u1_coord) (soft_u1) {};
\node[rectangle, scale=0.5, fill=black] at (dynamics01_coord) (dynamics01) {};
\node[rectangle, scale=0.5, fill=red] at (dynamics12_coord) (dynamics12) {};
\end{tikzpicture}}}
            & $\rightarrow$ &
             \adjustbox{valign=m}{\resizebox{1.1\width}{!}{\begin{tikzpicture}
\definecolor {dgreen} {rgb} { 0,0.5,0 };
\coordinate (x0_coord) at (0.0, 0);
\coordinate (x1_coord) at (1.2, 0);
\coordinate (x2_coord) at (2.4, 0);
\coordinate (u0_coord) at (0.6, -1);
\coordinate (u1_coord) at (1.7999999999999998, -1);
\coordinate (soft_x0_coord) at (0.48, 0.2);
\coordinate (soft_x1_coord) at (1.68, 0.2);
\coordinate (soft_u0_coord) at (0.12, -1.2);
\coordinate (soft_u1_coord) at (1.32, -1.2);
\coordinate (dynamics01_coord) at (0.6, -0.3333333333333333);
\coordinate (marg_x2_coord) at (1.5, -0.5);
\path[draw=black, line width=0.3pt] (x0_coord) -- (soft_x0_coord);
\path[draw=black, line width=0.3pt] (x1_coord) -- (soft_x1_coord);
\path[draw=black, line width=0.3pt] (u0_coord) -- (soft_u0_coord);
\path[draw=black, line width=0.3pt] (u1_coord) -- (soft_u1_coord);
\path[draw=black, line width=0.3pt] (x0_coord) -- (dynamics01_coord);
\path[draw=black, line width=0.3pt] (u0_coord) -- (dynamics01_coord);
\path[draw=black, line width=0.3pt] (x1_coord) -- (dynamics01_coord);
\path[draw=blue, line width=0.3pt] (x1_coord) -- (marg_x2_coord);
\path[draw=blue, line width=0.3pt] (u1_coord) -- (marg_x2_coord);
\node[scale=0.7, fill=white][circle, inner sep=2.8pt, draw, very thin] at (x0_coord) (x0) {$x_0$};
\node[scale=0.7, fill=white][circle, inner sep=2.8pt, draw, very thin] at (x1_coord) (x1) {$x_1$};
\node[scale=0.7, fill=white][circle, inner sep=2.8pt, draw, very thin] at (x2_coord) (x2) {$x_2$};
\node[scale=0.7, fill=white][circle, inner sep=2.8pt, draw, very thin] at (u0_coord) (u0) {$u_0$};
\node[scale=0.7, fill=white][circle, inner sep=2.8pt, draw, very thin] at (u1_coord) (u1) {$u_1$};
\node[circle, scale=0.5, fill=black] at (soft_x0_coord) (soft_x0) {};
\node[circle, scale=0.5, fill=black] at (soft_x1_coord) (soft_x1) {};
\node[circle, scale=0.5, fill=black] at (soft_u0_coord) (soft_u0) {};
\node[circle, scale=0.5, fill=black] at (soft_u1_coord) (soft_u1) {};
\node[rectangle, scale=0.5, fill=black] at (dynamics01_coord) (dynamics01) {};
\node[circle, scale=0.5, fill=blue] at (marg_x2_coord) (marg_x2) {};
\tikzset{
  big green arrow/.style={
  decoration={markings,mark=at position 1 with {\arrow[scale=2,dgreen, >=latex]{>}}},
    postaction={decorate}}
}
\path[big green arrow, draw=dgreen, line width=0.3pt] (x1) -- (x2);
\path[big green arrow, draw=dgreen, line width=0.3pt] (u1) -- (x2);
\end{tikzpicture}}}
            \\[2.5em]
            \tabdashline & & \tabdashline\\[-0.75em]
            {\footnotesize
            $\begin{aligned}
                x_2^*(x_1, u_1) = \argmin_{x_2} \red{x_2^TQ_{xx_T}x_2} \\
                        \; \suchthat \red{x_2 = F_{x_1} x_1 + F_{u_1} u_1} \\
            \end{aligned}$
            }
            & $\rightarrow$ &
            {\footnotesize
            $\begin{aligned}
                x_2^*(u_1, x_1) &= \green{F_{x_1} x_1} + \green{F_{u_1} u_1} \\
                \phi_{x_2}^*(u_1, x_1) &= \blue{x_2^{*T} Q_{xx_T} x_2^*}
            \end{aligned}$
            }%
            \\[1.7em]
            \tabdashline & & \tabdashline\\[-0.5em]
            \footnotesize
            $
                \stackrel{
                \hspace{1pt} W_{x_2} \hspace{13pt} x_2 \hspace{20pt} u_1 \hspace{20pt} x_1 \hfill}{
                \begin{bmatrix} I \\ \infty \end{bmatrix}
                \left[\begin{array}{@{}lll|l@{}}
                \red{Q_{xx_T}^{1/2}} &    \red{0} &  \red{0}  &     \red{0}\\
                \red{I} & \red{-F_{u_1}}  & \red{-F_{x_1}}  &\red{0}
                \end{array}\right]}$
            & $\rightarrow$ &
            \footnotesize
            $\stackrel{
                \hspace{1pt} W_{x_2}' \hspace{10pt} x_2 \hspace{17pt} u_1 \hspace{28pt} x_1 \hfill}{
                \begin{bmatrix} \infty \\ I \end{bmatrix}
                \left[\begin{array}{@{}lll|l@{}}
                \green{I} & \green{-F_{u_1}}  & \green{-F_{x_1}}  &\green{0} \\
                0 & \blue{Q_{xx_T}^{\frac{1}{2}}F_{u_1}}  & \blue{Q_{xx_T}^{\frac{1}{2}}F_{x_1}}   &     \blue{0}\\
                \end{array}\right]}$\\[8pt]
        \end{tabular}
        \subcaption{Eliminate $x_2$} \label{fig:LQR_elimx2}
    \end{subfigure} 
    \unskip \hspace*{8pt}\vrule~
    \begin{subfigure}{0.47\linewidth}
    \vspace{0.4em}
        \centering
        \begin{tabular}{ccc}
            \adjustbox{valign=m}{\resizebox{1.1\width}{!}{\begin{tikzpicture}
\definecolor {dgreen} {rgb} { 0,0.5,0 };
\coordinate (x0_coord) at (0.0, 0);
\coordinate (x1_coord) at (1.2, 0);
\coordinate (x2_coord) at (2.4, 0);
\coordinate (u0_coord) at (0.6, -1);
\coordinate (u1_coord) at (1.7999999999999998, -1);
\coordinate (soft_x0_coord) at (0.48, 0.2);
\coordinate (soft_x1_coord) at (1.68, 0.2);
\coordinate (soft_u0_coord) at (0.12, -1.2);
\coordinate (soft_u1_coord) at (1.32, -1.2);
\coordinate (dynamics01_coord) at (0.6, -0.3333333333333333);
\coordinate (marg_x2_coord) at (1.5, -0.5);
\path[draw=black, line width=0.3pt] (x0_coord) -- (soft_x0_coord);
\path[draw=black, line width=0.3pt] (x1_coord) -- (soft_x1_coord);
\path[draw=black, line width=0.3pt] (u0_coord) -- (soft_u0_coord);
\path[draw=red, line width=0.3pt] (u1_coord) -- (soft_u1_coord);
\path[draw=black, line width=0.3pt] (x0_coord) -- (dynamics01_coord);
\path[draw=black, line width=0.3pt] (u0_coord) -- (dynamics01_coord);
\path[draw=black, line width=0.3pt] (x1_coord) -- (dynamics01_coord);
\path[draw=red, line width=0.3pt] (x1_coord) -- (marg_x2_coord);
\path[draw=red, line width=0.3pt] (u1_coord) -- (marg_x2_coord);
\node[scale=0.7, fill=white][circle, inner sep=2.8pt, draw, very thin] at (x0_coord) (x0) {$x_0$};
\node[scale=0.7, fill=white][circle, inner sep=2.8pt, draw, very thin] at (x1_coord) (x1) {$x_1$};
\node[scale=0.7, fill=white][circle, inner sep=2.8pt, draw, very thin] at (x2_coord) (x2) {$x_2$};
\node[scale=0.7, fill=white][circle, inner sep=2.8pt, draw, very thin] at (u0_coord) (u0) {$u_0$};
\node[scale=0.7, fill=white][circle, inner sep=2.8pt, draw, very thin] at (u1_coord) (u1) {$u_1$};
\node[circle, scale=0.5, fill=black] at (soft_x0_coord) (soft_x0) {};
\node[circle, scale=0.5, fill=black] at (soft_x1_coord) (soft_x1) {};
\node[circle, scale=0.5, fill=black] at (soft_u0_coord) (soft_u0) {};
\node[circle, scale=0.5, fill=red] at (soft_u1_coord) (soft_u1) {};
\node[rectangle, scale=0.5, fill=black] at (dynamics01_coord) (dynamics01) {};
\node[circle, scale=0.5, fill=red] at (marg_x2_coord) (marg_x2) {};
\tikzset{
  big gray arrow/.style={
  decoration={markings,mark=at position 1 with {\arrow[scale=2,gray!50, >=latex]{>}}},
    postaction={decorate}}
}
\path[big gray arrow, draw=gray!50, line width=0.3pt] (x1) -- (x2);
\path[big gray arrow, draw=gray!50, line width=0.3pt] (u1) -- (x2);
\end{tikzpicture}}}
            & $\rightarrow$ &
            \adjustbox{valign=m}{\resizebox{1.1\width}{!}{\begin{tikzpicture}
\definecolor {dgreen} {rgb} { 0,0.5,0 };
\coordinate (x0_coord) at (0.0, 0);
\coordinate (x1_coord) at (1.2, 0);
\coordinate (x2_coord) at (2.4, 0);
\coordinate (u0_coord) at (0.6, -1);
\coordinate (u1_coord) at (1.7999999999999998, -1);
\coordinate (soft_x0_coord) at (0.48, 0.2);
\coordinate (soft_x1_coord) at (1.68, 0.2);
\coordinate (soft_u0_coord) at (0.12, -1.2);
\coordinate (dynamics01_coord) at (0.6, -0.3333333333333333);
\coordinate (marg_u1_coord) at (1.2, 0.4);
\path[draw=black, line width=0.3pt] (x0_coord) -- (soft_x0_coord);
\path[draw=black, line width=0.3pt] (x1_coord) -- (soft_x1_coord);
\path[draw=black, line width=0.3pt] (u0_coord) -- (soft_u0_coord);
\path[draw=black, line width=0.3pt] (x0_coord) -- (dynamics01_coord);
\path[draw=black, line width=0.3pt] (u0_coord) -- (dynamics01_coord);
\path[draw=black, line width=0.3pt] (x1_coord) -- (dynamics01_coord);
\path[draw=blue, line width=0.3pt] (x1_coord) -- (marg_u1_coord);
\node[scale=0.7, fill=white][circle, inner sep=2.8pt, draw, very thin] at (x0_coord) (x0) {$x_0$};
\node[scale=0.7, fill=white][circle, inner sep=2.8pt, draw, very thin] at (x1_coord) (x1) {$x_1$};
\node[scale=0.7, fill=white][circle, inner sep=2.8pt, draw, very thin] at (x2_coord) (x2) {$x_2$};
\node[scale=0.7, fill=white][circle, inner sep=2.8pt, draw, very thin] at (u0_coord) (u0) {$u_0$};
\node[scale=0.7, fill=white][circle, inner sep=2.8pt, draw, very thin] at (u1_coord) (u1) {$u_1$};
\node[circle, scale=0.5, fill=black] at (soft_x0_coord) (soft_x0) {};
\node[circle, scale=0.5, fill=black] at (soft_x1_coord) (soft_x1) {};
\node[circle, scale=0.5, fill=black] at (soft_u0_coord) (soft_u0) {};
\node[rectangle, scale=0.5, fill=black] at (dynamics01_coord) (dynamics01) {};
\node[circle, scale=0.5, fill=blue] at (marg_u1_coord) (marg_u1) {};
\tikzset{
  big green arrow/.style={
  decoration={markings,mark=at position 1 with {\arrow[scale=2,dgreen, >=latex]{>}}},
    postaction={decorate}}
}
\tikzset{
  big gray arrow/.style={
  decoration={markings,mark=at position 1 with {\arrow[scale=2,gray!50, >=latex]{>}}},
    postaction={decorate}}
}
\path[big gray arrow, draw=gray!50, line width=0.3pt] (x1) -- (x2);
\path[big gray arrow, draw=gray!50, line width=0.3pt] (u1) -- (x2);
\path[big green arrow, draw=dgreen, line width=0.3pt] (x1) -- (u1);
\end{tikzpicture}}}
            \\[2.5em]
            \tabdashline & & \tabdashline\\[-0.75em]
            {\footnotesize
                $\begin{aligned}
                    u_1^*(x_1) = \argmin_{u_1} &\red{\phi_{x_2}^*(x_1, u_1)}\\[-0.5em]
                    &+ \red{u_1^TQ_{uu_1}u_1}
                \end{aligned}$
            }%
            & $\rightarrow$ &\hspace{0.5ex}
            {\footnotesize
                $\begin{aligned}
                    u_1^*(x_1) = &\green{-K_1 x_1} \\
                    \phi_{u_1}^*(x_1) = &\blue{(K_1x_1)^TQ_{uu_1}(K_1x_1)} \\
                    & + \blue{\phi_{x_2}^*(x_1, -K_1 x_1)}
                \end{aligned}$
            }%
            \\[1.5em]
            \tabdashline & & \tabdashline\\[-0.5em]
            \footnotesize
            $\stackrel{
                \hspace{1pt} W_{u_1} \hspace{15pt} u_1 \hspace{30pt} x_1 \hfill}{
                \begin{bmatrix} I \\ I \end{bmatrix}
                \left[\begin{array}{@{}cc|c@{}}
                \red{Q_{xx_T}^{\frac{1}{2}}F_{u_1}} & \red{Q_{xx_T}^{\frac{1}{2}}F_{x_1}} & \red{0} \\
                \red{Q_{uu_1}^{\frac{1}{2}}} & \red{0}  &    \red{0}\\
                \end{array}\right]}$
            & $\rightarrow$ &
            \footnotesize
            $\stackrel{
                \hspace{4pt} W_{u_1}' \hspace{7pt} u_1 \hspace{7pt} x_1 \hfill}{
                \begin{bmatrix} R_1 \\ I \end{bmatrix}
                \left[\begin{array}{@{}cc|c@{}}
                \green{I} & \green{K_1} & \green{0} \\
                 & \blue{E_1} & \blue{0} \\
                \end{array}\right]}$
        \end{tabular}
        \subcaption{Eliminate $u_1$} \label{fig:LQR_elimu1}
    \end{subfigure} \\
    
    \caption{\footnotesize{Two variable eliminations for the LQR problem.  Each sub-figure consists of three rows showing three equivalent representations: the factor graph (top), constrained optimization (middle), and modified Gram-Schmidt process on $[A_i|b_i]$ 
    (bottom). The arrows in the factor graphs show variable dependencies. The thin horizontal arrows separate cases before and after elimination. Terms and symbols in the same color correspond to the color-coded variable elimination steps in Section \ref{sec:lqr}.  Note that the matrix factorization representation consists of the weight vector, $W_{i}$, next to the sub-matrix $[A_i|b_i]$.}
    } \label{fig:lqr_VE}
\end{figure*}
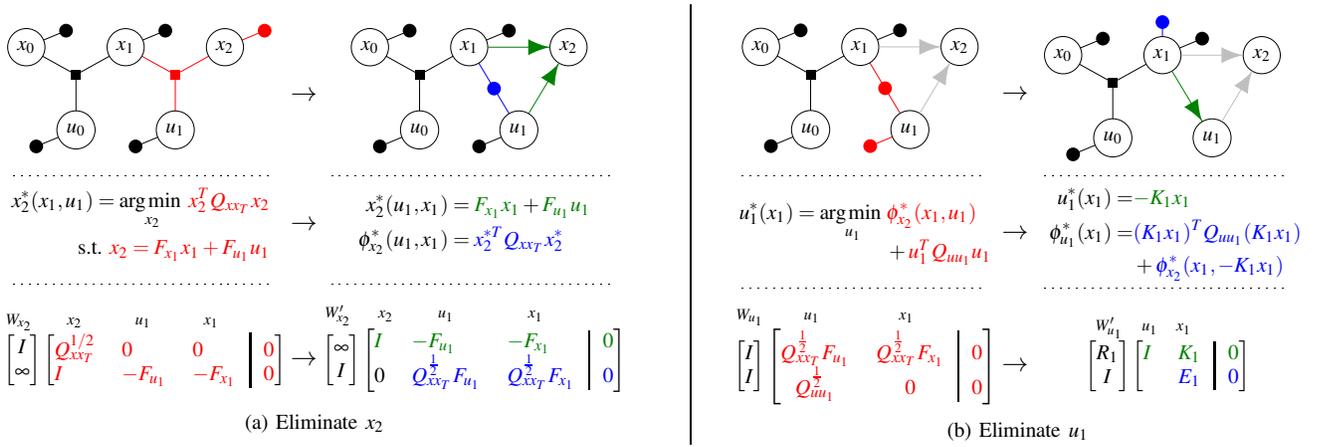
\begin{figure*}
    \setlength\tabcolsep{1pt}
    \hspace*{-10pt} 
    \begin{subfigure}[t]{0.56\linewidth}
        \centering
        \begin{tabular}{ccc}
            \adjustbox{valign=m}{\resizebox{1.2\width}{!}{\begin{tikzpicture}
\definecolor {dgreen} {rgb} { 0,0.5,0 };
\coordinate (x0_coord) at (0.0, 0);
\coordinate (x1_coord) at (1.2, 0);
\coordinate (x2_coord) at (2.4, 0);
\coordinate (u0_coord) at (0.6, -1);
\coordinate (u1_coord) at (1.7999999999999998, -1);
\coordinate (soft_x0_coord) at (0.48, 0.2);
\coordinate (soft_x1_coord) at (1.68, 0.2);
\coordinate (soft_x2_coord) at (2.88, 0.2);
\coordinate (soft_u0_coord) at (0.12, -1.2);
\coordinate (soft_u1_coord) at (1.32, -1.2);
\coordinate (dynamics01_coord) at (0.6, -0.3333333333333333);
\coordinate (dynamics12_coord) at (1.8, -0.3333333333333333);
\coordinate (constrain0_coord) at (0.3, -0.5);
\coordinate (constrain1_coord) at (1.5, -0.5);
\coordinate (constrain2_coord) at (2.6999999999999997, -0.5);
\path[draw=black, line width=0.3pt] (x0_coord) -- (soft_x0_coord);
\path[draw=black, line width=0.3pt] (x1_coord) -- (soft_x1_coord);
\path[draw=red, line width=0.3pt] (x2_coord) -- (soft_x2_coord);
\path[draw=black, line width=0.3pt] (u0_coord) -- (soft_u0_coord);
\path[draw=black, line width=0.3pt] (u1_coord) -- (soft_u1_coord);
\path[draw=black, line width=0.3pt] (x0_coord) -- (dynamics01_coord);
\path[draw=black, line width=0.3pt] (u0_coord) -- (dynamics01_coord);
\path[draw=black, line width=0.3pt] (x1_coord) -- (dynamics01_coord);
\path[draw=red, line width=0.3pt] (x1_coord) -- (dynamics12_coord);
\path[draw=red, line width=0.3pt] (u1_coord) -- (dynamics12_coord);
\path[draw=red, line width=0.3pt] (x2_coord) -- (dynamics12_coord);
\path[draw=black, line width=0.3pt] (x0_coord) -- (constrain0_coord);
\path[draw=black, line width=0.3pt] (u0_coord) -- (constrain0_coord);
\path[draw=black, line width=0.3pt] (x1_coord) -- (constrain1_coord);
\path[draw=black, line width=0.3pt] (u1_coord) -- (constrain1_coord);
\path[draw=red, line width=0.3pt] (x2_coord) -- (constrain2_coord);
\node[scale=0.7, fill=white][circle, inner sep=2.8pt, draw, very thin] at (x0_coord) (x0) {$x_0$};
\node[scale=0.7, fill=white][circle, inner sep=2.8pt, draw, very thin] at (x1_coord) (x1) {$x_1$};
\node[scale=0.7, fill=white][circle, inner sep=2.8pt, draw, very thin] at (x2_coord) (x2) {$x_2$};
\node[scale=0.7, fill=white][circle, inner sep=2.8pt, draw, very thin] at (u0_coord) (u0) {$u_0$};
\node[scale=0.7, fill=white][circle, inner sep=2.8pt, draw, very thin] at (u1_coord) (u1) {$u_1$};
\node[circle, scale=0.5, fill=black] at (soft_x0_coord) (soft_x0) {};
\node[circle, scale=0.5, fill=black] at (soft_x1_coord) (soft_x1) {};
\node[circle, scale=0.5, fill=red] at (soft_x2_coord) (soft_x2) {};
\node[circle, scale=0.5, fill=black] at (soft_u0_coord) (soft_u0) {};
\node[circle, scale=0.5, fill=black] at (soft_u1_coord) (soft_u1) {};
\node[rectangle, scale=0.5, fill=black] at (dynamics01_coord) (dynamics01) {};
\node[rectangle, scale=0.5, fill=red] at (dynamics12_coord) (dynamics12) {};
\node[rectangle, scale=0.5, fill=black] at (constrain0_coord) (constrain0) {};
\node[rectangle, scale=0.5, fill=black] at (constrain1_coord) (constrain1) {};
\node[rectangle, scale=0.5, fill=red] at (constrain2_coord) (constrain2) {};
\end{tikzpicture}}}
            & $\rightarrow$ &
            \adjustbox{valign=m}{\resizebox{1.2\width}{!}{\begin{tikzpicture}
\definecolor {dgreen} {rgb} { 0,0.5,0 };
\coordinate (x0_coord) at (0.0, 0);
\coordinate (x1_coord) at (1.2, 0);
\coordinate (x2_coord) at (2.4, 0);
\coordinate (u0_coord) at (0.6, -1);
\coordinate (u1_coord) at (1.7999999999999998, -1);
\coordinate (soft_x0_coord) at (0.48, 0.2);
\coordinate (soft_x1_coord) at (1.68, 0.2);
\coordinate (soft_u0_coord) at (0.12, -1.2);
\coordinate (soft_u1_coord) at (1.32, -1.2);
\coordinate (dynamics01_coord) at (0.6, -0.3333333333333333);
\coordinate (constrain0_coord) at (0.3, -0.5);
\coordinate (constrain1_coord) at (1.5, -0.5);
\coordinate (hard_marg_x2_coord) at (1.7999999999999998, -0.4);
\coordinate (soft_marg_x2_coord) at (1.2, -0.6);
\path[draw=black, line width=0.3pt] (x0_coord) -- (soft_x0_coord);
\path[draw=black, line width=0.3pt] (x1_coord) -- (soft_x1_coord);
\path[draw=black, line width=0.3pt] (u0_coord) -- (soft_u0_coord);
\path[draw=black, line width=0.3pt] (u1_coord) -- (soft_u1_coord);
\path[draw=black, line width=0.3pt] (x0_coord) -- (dynamics01_coord);
\path[draw=black, line width=0.3pt] (u0_coord) -- (dynamics01_coord);
\path[draw=black, line width=0.3pt] (x1_coord) -- (dynamics01_coord);
\path[draw=black, line width=0.3pt] (x0_coord) -- (constrain0_coord);
\path[draw=black, line width=0.3pt] (u0_coord) -- (constrain0_coord);
\path[draw=black, line width=0.3pt] (x1_coord) -- (constrain1_coord);
\path[draw=black, line width=0.3pt] (u1_coord) -- (constrain1_coord);
\path[draw=blue, line width=0.3pt] (x1_coord) -- (hard_marg_x2_coord);
\path[draw=blue, line width=0.3pt] (u1_coord) -- (hard_marg_x2_coord);
\path[draw=blue, line width=0.3pt] (x1_coord) -- (soft_marg_x2_coord);
\path[draw=blue, line width=0.3pt] (u1_coord) -- (soft_marg_x2_coord);
\node[scale=0.7, fill=white][circle, inner sep=2.8pt, draw, very thin] at (x0_coord) (x0) {$x_0$};
\node[scale=0.7, fill=white][circle, inner sep=2.8pt, draw, very thin] at (x1_coord) (x1) {$x_1$};
\node[scale=0.7, fill=white][circle, inner sep=2.8pt, draw, very thin] at (x2_coord) (x2) {$x_2$};
\node[scale=0.7, fill=white][circle, inner sep=2.8pt, draw, very thin] at (u0_coord) (u0) {$u_0$};
\node[scale=0.7, fill=white][circle, inner sep=2.8pt, draw, very thin] at (u1_coord) (u1) {$u_1$};
\node[circle, scale=0.5, fill=black] at (soft_x0_coord) (soft_x0) {};
\node[circle, scale=0.5, fill=black] at (soft_x1_coord) (soft_x1) {};
\node[circle, scale=0.5, fill=black] at (soft_u0_coord) (soft_u0) {};
\node[circle, scale=0.5, fill=black] at (soft_u1_coord) (soft_u1) {};
\node[rectangle, scale=0.5, fill=black] at (dynamics01_coord) (dynamics01) {};
\node[rectangle, scale=0.5, fill=black] at (constrain0_coord) (constrain0) {};
\node[rectangle, scale=0.5, fill=black] at (constrain1_coord) (constrain1) {};
\node[rectangle, scale=0.5, fill=blue] at (hard_marg_x2_coord) (hard_marg_x2) {};
\node[circle, scale=0.5, fill=blue] at (soft_marg_x2_coord) (soft_marg_x2) {};
\tikzset{
  big green arrow/.style={
  decoration={markings,mark=at position 1 with {\arrow[scale=2,dgreen, >=latex]{>}}},
    postaction={decorate}}
}
\path[big green arrow, draw=dgreen, line width=0.3pt] (x1) -- (x2);
\path[big green arrow, draw=dgreen, line width=0.3pt] (u1) -- (x2);
\end{tikzpicture}}}
            \\[2.5em]
            \tabdashline & & \tabdashline\\[-0.75em]
            {\footnotesize
            $\begin{aligned}
                x_2^*(x_1, u_1) = \argmin_{x_2} \red{x_2^TQ_{xx_T}x_2} \\
                        \; \suchthat \red{G_{x_2}x_2 - g_{l_2} = 0} \\
                        \phantom{\; \suchthat} \red{x_2 - F_{u_1}  u_1 - F_{x_1} x_1 = 0}
            \end{aligned}$
            }
            & $\rightarrow$ &
            {\footnotesize
            $\begin{aligned}
                x_2^*(u_1, x_1) &= \green{F_{x_1} x_1} + \green{F_{u_1}  u_1} \\
                \phi_{x_2}^*(u_1, x_1) &= \blue{||F_{x_1}x_1+F_{u_1} u_1||_{Q_{xx_T}}^2} \\
                \psi_{x_2}^*(u_1, x_1) &= \blue{G_{x_2}F_{u_1}u_1+G_{x_2}F_{x_1}x_1-g_{l_2}} = 0
            \end{aligned}$
            }%
            \\[2em]
            \tabdashline & & \tabdashline\\[-0.2em]
            \footnotesize
            $
                \stackrel{
                \hspace{5pt} W_{x_2} \hspace{15pt} x_2 \hspace{23pt} u_1 \hspace{20pt} x_1 \hfill}{\hspace{0.1em}
                \begin{bmatrix} I \\ \infty \\ \infty \end{bmatrix}
                \left[\begin{array}{@{}ccc|c@{}}
                \red{Q_{xx_T}^{\frac{1}{2}}} &   &    &     0\\
                \red{G_{x_2}} &     &   &  \red{g_{l_2}}\\
                \red{I} & \red{-F_{u_1}}  & \red{-F_{x_1}}  &0
                \end{array}\right]}$
            & $\rightarrow$ &
            \footnotesize
            $\stackrel{
                \hspace{5pt} W_{x_2}' \hspace{8pt} x_2 \hspace{19pt} u_1 \hspace{27pt} x_1 \hfill}{
                \begin{bmatrix} \infty \\ I \\ \infty \end{bmatrix}
                \left[\begin{array}{@{}ccc|c@{}}
                \green{I} & \green{-F_{u_1}}  & \green{-F_{x_1}}  &0 \\
                0 & \blue{Q_{xx_T}^{\frac{1}{2}}F_{u_1}}  & \blue{Q_{xx_T}^{\frac{1}{2}}F_{x_1}}   &     0\\
                0 & \blue{G_{x_2}F_{u_1}}    & \blue{G_{x_2}F_{x_1}}  &  \blue{g_{l_2}}\\
                \end{array}\right]}$ \\[1.75em]
        \end{tabular}
        \caption{Eliminate $x_2$} \label{fig:ecLQR_elimx2}
    \end{subfigure} 
    \unskip \hspace*{0pt}\vrule\hspace*{-3pt}~
    \begin{subfigure}[t]{0.43\linewidth}
        \centering
        \begin{tabular}{ccc}
            \adjustbox{valign=m}{\resizebox{1.2\width}{!}{\begin{tikzpicture}
\definecolor {dgreen} {rgb} { 0,0.5,0 };
\coordinate (x0_coord) at (0.0, 0);
\coordinate (x1_coord) at (1.2, 0);
\coordinate (x2_coord) at (2.4, 0);
\coordinate (u0_coord) at (0.6, -1);
\coordinate (u1_coord) at (1.7999999999999998, -1);
\coordinate (soft_x0_coord) at (0.48, 0.2);
\coordinate (soft_x1_coord) at (1.68, 0.2);
\coordinate (soft_u0_coord) at (0.12, -1.2);
\coordinate (soft_u1_coord) at (1.32, -1.2);
\coordinate (dynamics01_coord) at (0.6, -0.3333333333333333);
\coordinate (constrain0_coord) at (0.3, -0.5);
\coordinate (constrain1_coord) at (1.5, -0.5);
\coordinate (hard_marg_x2_coord) at (1.7999999999999998, -0.4);
\coordinate (soft_marg_x2_coord) at (1.2, -0.6);
\path[draw=black, line width=0.3pt] (x0_coord) -- (soft_x0_coord);
\path[draw=black, line width=0.3pt] (x1_coord) -- (soft_x1_coord);
\path[draw=black, line width=0.3pt] (u0_coord) -- (soft_u0_coord);
\path[draw=red, line width=0.3pt] (u1_coord) -- (soft_u1_coord);
\path[draw=black, line width=0.3pt] (x0_coord) -- (dynamics01_coord);
\path[draw=black, line width=0.3pt] (u0_coord) -- (dynamics01_coord);
\path[draw=black, line width=0.3pt] (x1_coord) -- (dynamics01_coord);
\path[draw=black, line width=0.3pt] (x0_coord) -- (constrain0_coord);
\path[draw=black, line width=0.3pt] (u0_coord) -- (constrain0_coord);
\path[draw=red, line width=0.3pt] (x1_coord) -- (constrain1_coord);
\path[draw=red, line width=0.3pt] (u1_coord) -- (constrain1_coord);
\path[draw=red, line width=0.3pt] (x1_coord) -- (hard_marg_x2_coord);
\path[draw=red, line width=0.3pt] (u1_coord) -- (hard_marg_x2_coord);
\path[draw=red, line width=0.3pt] (x1_coord) -- (soft_marg_x2_coord);
\path[draw=red, line width=0.3pt] (u1_coord) -- (soft_marg_x2_coord);
\node[scale=0.7, fill=white][circle, inner sep=2.8pt, draw, very thin] at (x0_coord) (x0) {$x_0$};
\node[scale=0.7, fill=white][circle, inner sep=2.8pt, draw, very thin] at (x1_coord) (x1) {$x_1$};
\node[scale=0.7, fill=white][circle, inner sep=2.8pt, draw, very thin] at (x2_coord) (x2) {$x_2$};
\node[scale=0.7, fill=white][circle, inner sep=2.8pt, draw, very thin] at (u0_coord) (u0) {$u_0$};
\node[scale=0.7, fill=white][circle, inner sep=2.8pt, draw, very thin] at (u1_coord) (u1) {$u_1$};
\node[circle, scale=0.5, fill=black] at (soft_x0_coord) (soft_x0) {};
\node[circle, scale=0.5, fill=black] at (soft_x1_coord) (soft_x1) {};
\node[circle, scale=0.5, fill=black] at (soft_u0_coord) (soft_u0) {};
\node[circle, scale=0.5, fill=red] at (soft_u1_coord) (soft_u1) {};
\node[rectangle, scale=0.5, fill=black] at (dynamics01_coord) (dynamics01) {};
\node[rectangle, scale=0.5, fill=black] at (constrain0_coord) (constrain0) {};
\node[rectangle, scale=0.5, fill=red] at (constrain1_coord) (constrain1) {};
\node[rectangle, scale=0.5, fill=red] at (hard_marg_x2_coord) (hard_marg_x2) {};
\node[circle, scale=0.5, fill=red] at (soft_marg_x2_coord) (soft_marg_x2) {};

\tikzset{
  big gray arrow/.style={
  decoration={markings,mark=at position 1 with {\arrow[scale=2,gray!50, >=latex]{>}}},
    postaction={decorate}}
}

\path[big gray arrow, draw=gray!50, line width=0.3pt] (x1) -- (x2);
\path[big gray arrow, draw=gray!50, line width=0.3pt] (u1) -- (x2);

\end{tikzpicture}}}
            & $\rightarrow$ &
            \adjustbox{valign=m}{\resizebox{1.2\width}{!}{\begin{tikzpicture}
\definecolor {dgreen} {rgb} { 0,0.5,0 };
\coordinate (x0_coord) at (0.0, 0);
\coordinate (x1_coord) at (1.2, 0);
\coordinate (x2_coord) at (2.1, 0);
\coordinate (u0_coord) at (0.6, -1);
\coordinate (u1_coord) at (1.7999999999999998, -1);
\coordinate (soft_x0_coord) at (0.48, 0.2);
\coordinate (soft_x1_coord) at (1.68, 0.2);
\coordinate (soft_u0_coord) at (0.12, -1.2);
\coordinate (dynamics01_coord) at (0.6, -0.3333333333333333);
\coordinate (constrain0_coord) at (0.3, -0.5);
\coordinate (soft_marg_u1_coord) at (1.2, 0.4);
\coordinate (hard_marg_u1_coord) at (0.96, 0.3);
\path[draw=black, line width=0.3pt] (x0_coord) -- (soft_x0_coord);
\path[draw=black, line width=0.3pt] (x1_coord) -- (soft_x1_coord);
\path[draw=black, line width=0.3pt] (u0_coord) -- (soft_u0_coord);
\path[draw=black, line width=0.3pt] (x0_coord) -- (dynamics01_coord);
\path[draw=black, line width=0.3pt] (u0_coord) -- (dynamics01_coord);
\path[draw=black, line width=0.3pt] (x1_coord) -- (dynamics01_coord);
\path[draw=black, line width=0.3pt] (x0_coord) -- (constrain0_coord);
\path[draw=black, line width=0.3pt] (u0_coord) -- (constrain0_coord);
\path[draw=blue, line width=0.3pt] (x1_coord) -- (soft_marg_u1_coord);
\path[draw=blue, line width=0.3pt] (x1_coord) -- (hard_marg_u1_coord);
\node[scale=0.7, fill=white][circle, inner sep=2.8pt, draw, very thin] at (x0_coord) (x0) {$x_0$};
\node[scale=0.7, fill=white][circle, inner sep=2.8pt, draw, very thin] at (x1_coord) (x1) {$x_1$};
\node[scale=0.7, fill=white][circle, inner sep=2.8pt, draw, very thin] at (x2_coord) (x2) {$x_2$};
\node[scale=0.7, fill=white][circle, inner sep=2.8pt, draw, very thin] at (u0_coord) (u0) {$u_0$};
\node[scale=0.7, fill=white][circle, inner sep=2.8pt, draw, very thin] at (u1_coord) (u1) {$u_1$};
\node[circle, scale=0.5, fill=black] at (soft_x0_coord) (soft_x0) {};
\node[circle, scale=0.5, fill=black] at (soft_x1_coord) (soft_x1) {};
\node[circle, scale=0.5, fill=black] at (soft_u0_coord) (soft_u0) {};
\node[rectangle, scale=0.5, fill=black] at (dynamics01_coord) (dynamics01) {};
\node[rectangle, scale=0.5, fill=black] at (constrain0_coord) (constrain0) {};
\node[circle, scale=0.5, fill=blue] at (soft_marg_u1_coord) (soft_marg_u1) {};
\node[rectangle, scale=0.5, fill=blue] at (hard_marg_u1_coord) (hard_marg_u1) {};

\tikzset{
  big gray arrow/.style={
  decoration={markings,mark=at position 1 with {\arrow[scale=2,gray!50, >=latex]{>}}},
    postaction={decorate}}
}

\path[big gray arrow, draw=gray!50, line width=0.3pt] (x1) -- (x2);
\path[big gray arrow, draw=gray!50, line width=0.3pt] (u1) -- (x2);
\tikzset{
  big green arrow/.style={
  decoration={markings,mark=at position 1 with {\arrow[scale=2,dgreen, >=latex]{>}}},
    postaction={decorate}}
}
\path[big green arrow, draw=dgreen, line width=0.3pt] (x1) -- (u1);
\end{tikzpicture}}}
            \\[2.5em]
            \tabdashline & & \tabdashline\\[-0.85em]
            {\footnotesize
                $\begin{aligned}
                    u_1^*(x_1) = \argmin_{u_1} \red{\phi_{x_2}^*(x_1, u_1)} + \red{u_1^TRu_1}\\
                        \; \suchthat \red{G_{x_2}F_{u_1}u_1 - G_{x_2}F_{x_1}x_1 - g_{l_2} = 0} \\
                        \phantom{\; \suchthat} \red{G_{u_1}u_1 - G_{x_1} x_1 - g_{l_1} = 0}
                \end{aligned}$
            }%
            & $\rightarrow$ &
            {\footnotesize
                $\begin{aligned}
                    u_1^*(x_1) &= \green{-K_1 x_1 +k_1} \\
                    \phi_{u_1}^*(x_1) &= \blue{||P^{\frac{1}{2}}_1x_1 - p_1||^2}\\
                    \psi_{u_1}^*(x_1) &= \blue{H_1x_1 - h_1} = 0
                \end{aligned}$
            }%
            \\[2em]
            \tabdashline & & \tabdashline\\[-0.5em]
            \footnotesize
            $\stackrel{
                \hspace{4pt} W_{u_1} \hspace{18pt} u_1 \hspace{30pt} x_1 \hfill}{
                \begin{bmatrix} I \\ \infty \\ I \\ \infty \end{bmatrix}
                \left[\begin{array}{@{}cc|c@{}}
                \red{Q_{xx_T}^{\frac{1}{2}}F_{u_1}}  & \red{Q_{xx_T}^{\frac{1}{2}}F_{x_1}}   &     0\\
                \red{G_{x_2}F_{u_1}}    & \red{G_{x_2}F_{x_1}}  &  \red{g_{l_2}}\\
                \red{Q_{uu_1}^{\frac{1}{2}}} & & 0 \\
                \red{G_{u_1}}    & \red{G_{x_1}}  &  \red{g_{l_1}}\\
                \end{array}\right]}$
            & $\rightarrow$ &
            \footnotesize
            $\stackrel{
                \hspace{4pt} W_{u_1}' \hspace{11pt} u_1 \hspace{12pt} x_1 \hfill}{
                \begin{bmatrix} R_1 \\ \infty \\ I \end{bmatrix}
                \left[\begin{array}{@{}cc|c@{}}
                \green{I} & \green{K_1} & \green{k_1} \\
                0 & \blue{H_1} & \blue{h_1} \\
                0 & \blue{P^{\frac{1}{2}}_1} & \blue{p_1}
                \end{array}\right]}$
        \end{tabular}
        \caption{Eliminate $u_1$} \label{fig:ecLQR_elimu1}
    \end{subfigure} \\
    \caption{\footnotesize{Two elimination steps for EC-LQR with local constraints. This figure has the same layout as Figure \ref{fig:lqr_VE}.}} \label{fig:eclqr_VE}
\end{figure*}
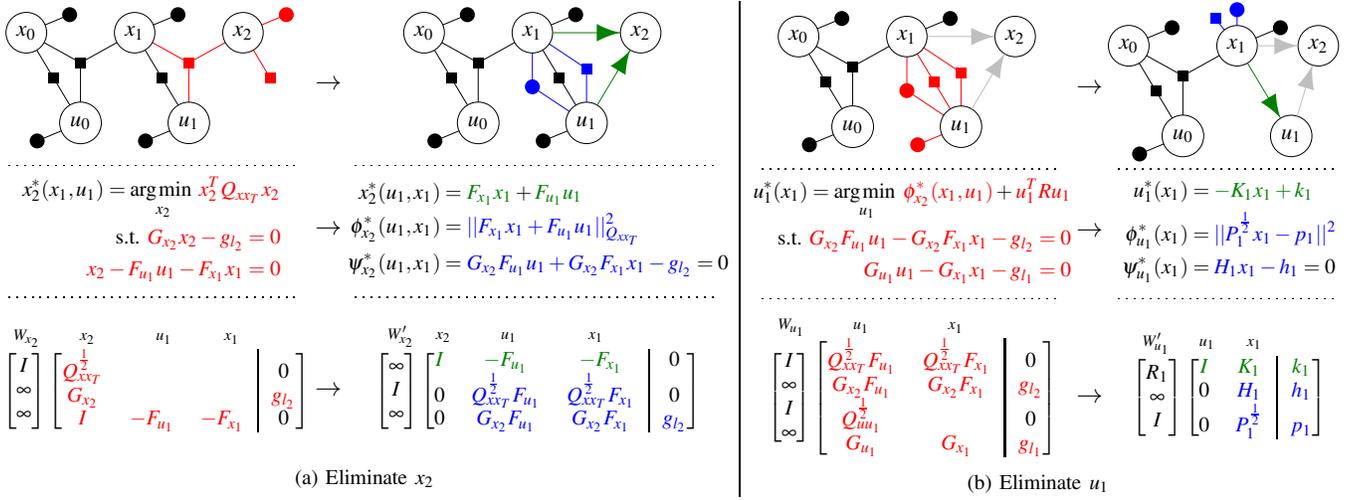



We first demonstrate how to represent standard LQR, Problem \ref{problem1} with only constraint (\ref{eqn:linear_sys_dyn}), as the factor graph shown in Fig. \ref{fig:lqr_suba} and subsequently obtain the optimal trajectory and optimal feedback control policy using VE.


Factor graphs can be interpreted as describing either a joint probability distribution with conditional independencies or, as we focus on in this paper, an equivalent least-squares problem derived from minimizing the negative log-likelihood.  A factor graph is a bipartite graph consisting of variables and factors connected by edges, where a factor can be viewed either as a joint probability density or least squares objective over the variables it is connected to.

We begin by showing how the probabilistic view of factor graphs is equivalent to a least squares minimization \cite{dellaert2017factor}.
We construct factor graph to describe a joint probability distribution of the variables $X=[\textbf{x};\textbf{u}]$.
For Gaussian distributions, the probability distribution for a single objective or constraint factor $\phi_k$ can be written in matrix form as\vspace*{-.5em}
$$
\phi_k(X_k) \propto \exp \left\{ -\tfrac{1}{2}\| A_kX_k - b_k\|^2_{\Sigma_k}    \right\} \vspace{-.5em}
$$
where $\exp$ is the exponential function and $X_k$ contains the variables connected to the factor. $A_k$ and $b_k$ are a matrix and a vector with problem-specific values, $\Sigma_k$ is the covariance of the probability distribution, and $||\cdot||_\Sigma^2 \coloneqq (\cdot)^T\Sigma^{-1} (\cdot)$ denotes the square of the Mahalanobis norm. $A_k$, $b_k$, and $\Sigma_k$ together define the probability density of the factor. 

The product of all factors is the posterior distribution of $X$ whose MAP estimate solves the least squares problem \cite{dellaert2017factor}:
\begin{align}
&X^{MAP}  = \argmax_{X} \phi(X) = \argmin_{X} -\log(\prod_k \phi_k(X_k)) \nonumber \\
& \; = \argmin_{X} \sum_k \| A_kX_k - b_k\|^2_{\Sigma_k}  = \argmin_{X} \| AX - b\|^2_{\Sigma}  \label{eqn:Ax=b}
\end{align}
{where $A$ and $\Sigma$ contain $A_k$ and $\Sigma_k$ on the block diagonal respectively and $b$ stacks all $b_k$ vertically. In this formulation, each factor $\phi_k$ corresponds to a block row in $[A|b]$.} Defining the weight matrix $W\coloneqq\Sigma^{-1}$, $X^{MAP}$ minimizes a weighted least squares expression $(AX - b)^TW(AX - b)$.

The objective factors in Fig. \ref{fig:lqr_suba} are
  $\phi_{objx} (x_t)\propto  \exp \{ -\frac{1}{2}\| Q^{1/2}_{xx_t}x_t \|^2    \}$ and
  $\phi_{obju} (u_t) \propto \exp \{ -\frac{1}{2}\| Q^{1/2}_{uu_t}u_t \|^2   \}$, while the constraint factors are 
  $
  \phi_{dyn} (x_{t+1}, x_t, u_t)  \propto \exp \{ -\frac{1}{2}\| x_{t+1} - F_{x_t}x_t - F_{u_t}u_t  \|^2_{\Sigma_c}    \}
  $ where the covariance $\Sigma_c$ = 0 creates infinite terms in $W$.
When factor graphs have factors with zero covariance, the least squares problem turns into a \textit{constrained} least squares problem which we can solve using e.g. modified Gram-Schmidt \cite{gulliksson1995modified}. If linear terms are desired in the cost function in (\ref{eqn:linear_cost}) (e.g. track a non-zero setpoint), we can always express the objective factor in a Gaussian form as $\phi_{objx} (x_t)\propto  \exp \{ -\frac{1}{2}\| Q^{1/2}_{xx_t}(x_t-x_{ref}) \|^2    \}$, where $x_{ref}$ is some tracking target.
  
The VE algorithm 
is a method to solve (\ref{eqn:Ax=b}) while exploiting the sparsity of $A$ by solving for one variable at a time. For a variable $\theta_i \in X$, we can identify its \textit{separator} $S_i$: the set of other variables sharing factors with $\theta_i$. Then we extract sub-matrices $A_i$, $W_i$, and sub-vector $b_i$ from the rows of $A$, $W$, and $b$ such that $[A_i|b_i]$ contains all factors connected to $\theta_i$. We collect the rows in $[A_i|b_i]$ with finite weights to define objective factor $\phi_i(\theta_i, S_i)$ and rows with infinite weights to define constraint factor $\psi_i(\theta_i, S_i)$.  
Then we ``eliminate'' variable $\theta_i$ following 3 steps\footnote{In the probabilistic form, steps \green{2} and \blue{3} would come from factoring $\phi_i(\theta_i, S_i)\psi_i(\theta_i, S_i) \propto p(\theta_i|S_i)p(S_i)$. For Gaussian distributions, $\theta_i^*(S_i) = E[p(\theta_i|S_i)]$ and $\phi_i^*(S_i)\psi_i^*(S_i) = p(S_i)$.}:
\begin{itemize}[leftmargin=12mm]
    \item[\textcolor{red}{Step 1.}]  Identify all the factors adjacent to $\theta_i$ to get $[A_i|b_i]$.  Split $[A_i|b_i]$ into  $\phi_i(\theta_i, S_i)$ and $\psi_i(\theta_i, S_i)$.
    \item[\textcolor{green}{Step 2.}] Solve the (constrained) least squares problem:\vspace*{-0.75ex}
    $$\theta_i^*(S_i) = \displaystyle \argmin_{\theta_i} \phi_i(\theta_i, S_i) \; \suchthat \psi_i(\theta_i, S_i) = 0$$
    using modified Gram-Schmidt or other constrained optimization methods \cite[Ch.10]{boyd2004convex}. $\theta_i^*(S_i)$ denotes that $\theta_i^*$ is a function of the variables in $S_i$. 
    
    \item[\textcolor{blue}{Step 3.}] Substitute $\theta_i\leftarrow \theta_i^*$ by replacing the factors  $\phi_i(\theta_i, S_i)$ and $\psi_i(\theta_i, S_i)$ with $\phi_i^*(S_i) \coloneqq \phi_i(\theta_i^*, S_i)$ and $\psi_i^*(S_i) \coloneqq \psi_i(\theta_i^*, S_i)$, respectively, in $[A|b]$.
\end{itemize}

We follow an \textit{elimination order} \cite{koller2009probabilistic} to eliminate one variable $\theta_i \in X$ at a time. After all variables are eliminated, the factor matrix $A$ is effectively converted into an upper-triangular matrix $R$ allowing $X$ to be solved by matrix back-substitution. Therefore, one interpretation of the VE algorithm is performing sparse QR factorization on $A$ \cite{dellaert2017factor}.



To apply VE to the LQR factor graph in Fig. \ref{fig:lqr_suba}, we choose the ordering $x_N, u_{N-1}, x_{N-1}, \ldots, x_0$ and execute Steps 1-3 to eliminate each variable. This order is chosen to generate feedback policies where the controls are functions of the present states.
When eliminating a state $x_i$ for the special case of LQR, 
the constrained least-squares problem in \green{Step 2} is trivially solved as $x_i^*(u_{i-1}, x_{i-1}) = F_{u}u_{i-1} + F_{x}x_{i-1}$.  Additionally, $\psi_{x_i}^*$ will be empty since $\psi_{x_i}(x_i^*, u_{i-1}, x_{i-1})$ is satisfied for any choice of $u_{i-1}$ and $x_{i-1}$.  Fig. \ref{fig:LQR_elimx2} shows the factor graphs, corresponding optimization problems, and sub-matrices $[W_i][A_i|b_i]$ before and after eliminating $x_2$.

The optimal feedback control policy emerges when eliminating a control $u_i$. The combined constraint factor $\psi_{u_i}$ is empty (since $\psi_{x_{i+1}}^*$ is empty), so \green{Step 2} reduces to an unconstrained minimization problem. To solve it using QR factorization, split the objective $||A_i[u;x]||_2^2 = ||R_iu+T_ix||_2^2 + ||E_ix||_2^2$ using the QR factorization $A_i=Q\mbox{\scriptsize $\begin{bmatrix}R_i&T_i\\0&E_i\end{bmatrix}$}$ noting that $Q$ is orthogonal and thus doesn't change the norm.  Then, $u_i^*(x_i)=-K_ix_i$ where $K_i\coloneqq R_i^{-1}T_i$ efficiently optimizes the first term and $\phi_{u_i}^*(x_i)=||E_ix||_2^2$ is the new factor on $x$.  The elimination is shown in Fig. \ref{fig:LQR_elimu1}.

Furthermore, the \costtogo{} (or ``value function'' \cite{bertsekas1995dynamic}), which commonly appears in DP-based LQR literature, is visually evident in the (right) factor graph from Fig. \ref{fig:LQR_elimu1} as the sum of the two unary factors on $x_1$:
$$\costtogo{}_{1}(x_1) = x_1^TQx_1 + \blue{x_1^TE_1^2x_1}.$$
Continuing to eliminate the rest of the variables reveals the general formula of the \costtogo{} after applying block-QR elimination to solve for $K_i$ and $E_i$:
$$\costtogo{}_{i}(x_i) = x_i^T(Q_{xx_t} + F_{x_t}^TV_{i+1} F_{x_t} - K_i^T F_{u_i}^TV_{i+1}F_{x_t})x_i$$
where $V_{i+1}$ comes from
$\phi_{u_i}^*(x_i) + x_i^TQ_{xx_t}x_i = x_i^TV_ix_i$.

\subsection{EC-LQR with \Local Constraints} \label{sec:ecLQR}


The factor graph representation of EC-LQR with only \local constraints (\ref{eqn:local_cst}) and (\ref{eqn:local_cst_final}) in Problem \ref{problem1} is the same as the factor graph in Figure \ref{fig:cross-ec-LQR-factor-graph} but without the red square marked ``cross-time-step constraint''. 
We still use the same elimination order: ${x_2,u_1,x_1,u_0,x_0}$ to execute VE.

\subsubsection{Eliminating a state}
The process for eliminating a state involves one more constraint when generating $\psi_{x_i}^*(S_{x_i})$, but solving for $x_i$ remains the same as in standard LQR case.  Figure \ref{fig:ecLQR_elimx2} shows the process of eliminating $x_2$.

\subsubsection{Eliminating a control}
The process for eliminating a control is a constrained minimization with some constraints on $u_i$ derived from $\psi_{x_{i+1}}^*(u_i, x_i)$ and/or $G_{x_i}x_i + G_{u_i}u_i +g_{l_i}= 0$. The elimination procedure 
is shown in Figure \ref{fig:ecLQR_elimu1}.
%
From the result of eliminating $u_1$ as shown on the right in Figure \ref{fig:ecLQR_elimu1}, we observe that
\begin{itemize}
    \item the optimal control policy \green{$u_1^*(x_1) = -K_1x_1 + k_1$} falls out, 
    \item \blue{$\phi_{u_1}^*(x_1) = ||P_1^{1/2}x_1-p_1||^2$} corresponds to the $\costtogo{}(x_1) = x_1^TV_1x_1-v_1x_1$ from \cite{laine2019efficient} where $V_1 = P_1 + Q_{xx_1}$ and $v_1 = 2p_1^TP_1$, and
    \item \blue{$\psi_{u_1}^* = H_1x_1-h_1 = 0$} corresponds to the $\constrainttogo{}(x_1) = H_1x_1-h_1 = 0$ from \cite{laine2019efficient}
\end{itemize}


We continue with VE to eliminate the remaining variables similarly.  After each $u_i$ is eliminated, we can obtain an optimal control policy, \textit{constraint\_to\_go}, and \textit{cost\_to\_go} -- all of which being functions of $x_i$. 
When the problem is linear and all matrices are invertible or full column rank, the optimal solution is unique. We will demonstrate our method finding the unique optimal solution in Section \ref{sec:experiment}.



\subsection{Computational Complexity Analysis}\label{sec:computation}
Because \red{Step 1} collects only the factors connected to the variable we seek to eliminate, VE is very efficient and the complexity of eliminating a single variable is independent of the trajectory length.
When eliminating one variable, we factorize a matrix, $A_i$, whose rows consist of all the factors connected to the variable and whose columns correspond to the variable and its separator.  Thus, the maximum dimension of $A_i$ in EC-LQR problem with just \local constraints is $3n \times (2n+m)$ when eliminating a state or $(2n+m) \times (n+m)$ when eliminating a control. In the worst case, the QR factorization on this matrix has complexity $O(2(3n)(2n+m)^2) = O(24n^3+24n^2m+6nm^2)$ when eliminating a state or $O(2(2n+m)^2(n+m)) = O(8n^3+16n^2m+10nm^2+2m^3)$ when eliminating a control.  To obtain the solution from the sparse QR factorization result of $A$, we apply back substitution whose computation complexity is ~$O(n^2+m^2)$, so the overall computation complexity of solving the trajectory with length $T$ is $O(T\cdot(\kappa_1n^3+\kappa_2n^2m+\kappa_3nm^2+\kappa_4m^3))$, which is the same as the state of the art DP approach \cite{laine2019efficient}.

\subsection{EC-LQR with Cross-time-step Constraints} \label{sec:crosstimestepLQR}
\begin{figure}
    \centering
    \begin{subfigure}{\linewidth}
    \vspace{0.3em}
    \centering
        \resizebox{0.9\linewidth}{!}{\begin{tikzpicture}
\coordinate (x0_coord) at (0.0, 0);
\coordinate (x1_coord) at (1.2, 0);
\coordinate (x2_coord) at (2.4, 0);
\coordinate (x3_coord) at (3.5999999999999996, 0);
\coordinate (x4_coord) at (4.8, 0);
\coordinate (x5_coord) at (6.0, 0);
\coordinate (x6_coord) at (7.199999999999999, 0);
\coordinate (x7_coord) at (8.4, 0);
\coordinate (u0_coord) at (0.6, -1);
\coordinate (u1_coord) at (1.7999999999999998, -1);
\coordinate (u2_coord) at (3.0, -1);
\coordinate (u3_coord) at (4.2, -1);
\coordinate (u4_coord) at (5.3999999999999995, -1);
\coordinate (u5_coord) at (6.6, -1);
\coordinate (u6_coord) at (7.8, -1);
\coordinate (soft_x0_coord) at (0.0, 0.6);
\coordinate (soft_x1_coord) at (1.2, 0.6);
\coordinate (soft_x2_coord) at (2.4, 0.6);
\coordinate (soft_x3_coord) at (3.5999999999999996, 0.6);
\coordinate (soft_x4_coord) at (4.8, 0.6);
\coordinate (soft_x5_coord) at (6.0, 0.6);
\coordinate (soft_x6_coord) at (7.199999999999999, 0.6);
\coordinate (soft_x7_coord) at (8.4, 0.6);
\coordinate (dynamics0_coord) at (0.6, -0.3333333333333333);
\coordinate (soft_u0_coord) at (0.6, -1.6);
\coordinate (constrain0_coord) at (0.3, -0.5);
\coordinate (dynamics1_coord) at (1.7999999999999998, -0.3333333333333333);
\coordinate (soft_u1_coord) at (1.7999999999999998, -1.6);
\coordinate (constrain1_coord) at (1.5, -0.5);
\coordinate (dynamics2_coord) at (3.0, -0.3333333333333333);
\coordinate (soft_u2_coord) at (3.0, -1.6);
\coordinate (constrain2_coord) at (2.7, -0.5);
\coordinate (dynamics3_coord) at (4.199999999999999, -0.3333333333333333);
\coordinate (soft_u3_coord) at (4.2, -1.6);
\coordinate (constrain3_coord) at (3.9, -0.5);
\coordinate (dynamics4_coord) at (5.3999999999999995, -0.3333333333333333);
\coordinate (soft_u4_coord) at (5.3999999999999995, -1.6);
\coordinate (constrain4_coord) at (5.1, -0.5);
\coordinate (dynamics5_coord) at (6.599999999999999, -0.3333333333333333);
\coordinate (soft_u5_coord) at (6.6, -1.6);
\coordinate (constrain5_coord) at (6.3, -0.5);
\coordinate (dynamics6_coord) at (7.8, -0.3333333333333333);
\coordinate (soft_u6_coord) at (7.8, -1.6);
\coordinate (constrain6_coord) at (7.5, -0.5);
\coordinate (cross1_coord) at (1.7999999999999998, 1);
\coordinate (cross2_coord) at (5.3999999999999995, 1);
\path[draw=black, line width=0.3pt] (x0_coord) -- (soft_x0_coord);
\path[draw=black, line width=0.3pt] (x1_coord) -- (soft_x1_coord);
\path[draw=black, line width=0.3pt] (x2_coord) -- (soft_x2_coord);
\path[draw=black, line width=0.3pt] (x3_coord) -- (soft_x3_coord);
\path[draw=black, line width=0.3pt] (x4_coord) -- (soft_x4_coord);
\path[draw=black, line width=0.3pt] (x5_coord) -- (soft_x5_coord);
\path[draw=black, line width=0.3pt] (x6_coord) -- (soft_x6_coord);
\path[draw=black, line width=0.3pt] (x7_coord) -- (soft_x7_coord);
\path[draw=black, line width=0.3pt] (x0_coord) -- (dynamics0_coord);
\path[draw=black, line width=0.3pt] (x1_coord) -- (dynamics0_coord);
\path[draw=black, line width=0.3pt] (u0_coord) -- (dynamics0_coord);
\path[draw=black, line width=0.3pt] (u0_coord) -- (soft_u0_coord);
\path[draw=black, line width=0.3pt] (x0_coord) -- (constrain0_coord);
\path[draw=black, line width=0.3pt] (u0_coord) -- (constrain0_coord);
\path[draw=black, line width=0.3pt] (x1_coord) -- (dynamics1_coord);
\path[draw=black, line width=0.3pt] (x2_coord) -- (dynamics1_coord);
\path[draw=black, line width=0.3pt] (u1_coord) -- (dynamics1_coord);
\path[draw=black, line width=0.3pt] (u1_coord) -- (soft_u1_coord);
\path[draw=black, line width=0.3pt] (x1_coord) -- (constrain1_coord);
\path[draw=black, line width=0.3pt] (u1_coord) -- (constrain1_coord);
\path[draw=black, line width=0.3pt] (x2_coord) -- (dynamics2_coord);
\path[draw=black, line width=0.3pt] (x3_coord) -- (dynamics2_coord);
\path[draw=black, line width=0.3pt] (u2_coord) -- (dynamics2_coord);
\path[draw=black, line width=0.3pt] (u2_coord) -- (soft_u2_coord);
\path[draw=black, line width=0.3pt] (x2_coord) -- (constrain2_coord);
\path[draw=black, line width=0.3pt] (u2_coord) -- (constrain2_coord);
\path[draw=black, line width=0.3pt] (x3_coord) -- (dynamics3_coord);
\path[draw=black, line width=0.3pt] (x4_coord) -- (dynamics3_coord);
\path[draw=black, line width=0.3pt] (u3_coord) -- (dynamics3_coord);
\path[draw=black, line width=0.3pt] (u3_coord) -- (soft_u3_coord);
\path[draw=black, line width=0.3pt] (x3_coord) -- (constrain3_coord);
\path[draw=black, line width=0.3pt] (u3_coord) -- (constrain3_coord);
\path[draw=black, line width=0.3pt] (x4_coord) -- (dynamics4_coord);
\path[draw=black, line width=0.3pt] (x5_coord) -- (dynamics4_coord);
\path[draw=black, line width=0.3pt] (u4_coord) -- (dynamics4_coord);
\path[draw=black, line width=0.3pt] (u4_coord) -- (soft_u4_coord);
\path[draw=black, line width=0.3pt] (x4_coord) -- (constrain4_coord);
\path[draw=black, line width=0.3pt] (u4_coord) -- (constrain4_coord);
\path[draw=black, line width=0.3pt] (x5_coord) -- (dynamics5_coord);
\path[draw=black, line width=0.3pt] (x6_coord) -- (dynamics5_coord);
\path[draw=black, line width=0.3pt] (u5_coord) -- (dynamics5_coord);
\path[draw=black, line width=0.3pt] (u5_coord) -- (soft_u5_coord);
\path[draw=black, line width=0.3pt] (x5_coord) -- (constrain5_coord);
\path[draw=black, line width=0.3pt] (u5_coord) -- (constrain5_coord);
\path[draw=black, line width=0.3pt] (x6_coord) -- (dynamics6_coord);
\path[draw=black, line width=0.3pt] (x7_coord) -- (dynamics6_coord);
\path[draw=black, line width=0.3pt] (u6_coord) -- (dynamics6_coord);
\path[draw=black, line width=0.3pt] (u6_coord) -- (soft_u6_coord);
\path[draw=black, line width=0.3pt] (x6_coord) -- (constrain6_coord);
\path[draw=black, line width=0.3pt] (u6_coord) -- (constrain6_coord);
\path[draw=red, line width=0.3pt] (x0_coord) -- (cross1_coord);
\path[draw=red, line width=0.3pt] (x3_coord) -- (cross1_coord);
\path[draw=red, line width=0.3pt] (x3_coord) -- (cross2_coord);
\path[draw=red, line width=0.3pt] (x6_coord) -- (cross2_coord);
\node[scale=0.8, fill=white][circle, inner sep=2.8pt, draw, very thin] at (x0_coord) (x0) {$x_0$};
\node[scale=0.8, fill=white][circle, inner sep=2.8pt, draw, very thin] at (x1_coord) (x1) {$x_1$};
\node[scale=0.8, fill=white][circle, inner sep=2.8pt, draw, very thin] at (x2_coord) (x2) {$x_2$};
\node[scale=0.8, fill=white][circle, inner sep=2.8pt, draw, very thin] at (x3_coord) (x3) {$x_3$};
\node[scale=0.8, fill=white][circle, inner sep=2.8pt, draw, very thin] at (x4_coord) (x4) {$x_4$};
\node[scale=0.8, fill=white][circle, inner sep=2.8pt, draw, very thin] at (x5_coord) (x5) {$x_5$};
\node[scale=0.8, fill=white][circle, inner sep=2.8pt, draw, very thin] at (x6_coord) (x6) {$x_6$};
\node[scale=0.8, fill=white][circle, inner sep=2.8pt, draw, very thin] at (x7_coord) (x7) {$x_7$};
\node[scale=0.8, fill=white][circle, inner sep=2.8pt, draw, very thin] at (u0_coord) (u0) {$u_0$};
\node[scale=0.8, fill=white][circle, inner sep=2.8pt, draw, very thin] at (u1_coord) (u1) {$u_1$};
\node[scale=0.8, fill=white][circle, inner sep=2.8pt, draw, very thin] at (u2_coord) (u2) {$u_2$};
\node[scale=0.8, fill=white][circle, inner sep=2.8pt, draw, very thin] at (u3_coord) (u3) {$u_3$};
\node[scale=0.8, fill=white][circle, inner sep=2.8pt, draw, very thin] at (u4_coord) (u4) {$u_4$};
\node[scale=0.8, fill=white][circle, inner sep=2.8pt, draw, very thin] at (u5_coord) (u5) {$u_5$};
\node[scale=0.8, fill=white][circle, inner sep=2.8pt, draw, very thin] at (u6_coord) (u6) {$u_6$};
\node[circle, scale=0.5, fill=black] at (soft_x0_coord) (soft_x0) {};
\node[circle, scale=0.5, fill=black] at (soft_x1_coord) (soft_x1) {};
\node[circle, scale=0.5, fill=black] at (soft_x2_coord) (soft_x2) {};
\node[circle, scale=0.5, fill=black] at (soft_x3_coord) (soft_x3) {};
\node[circle, scale=0.5, fill=black] at (soft_x4_coord) (soft_x4) {};
\node[circle, scale=0.5, fill=black] at (soft_x5_coord) (soft_x5) {};
\node[circle, scale=0.5, fill=black] at (soft_x6_coord) (soft_x6) {};
\node[circle, scale=0.5, fill=black] at (soft_x7_coord) (soft_x7) {};
\node[rectangle, scale=0.5, fill=black] at (dynamics0_coord) (dynamics0) {};
\node[circle, scale=0.5, fill=black] at (soft_u0_coord) (soft_u0) {};
\node[rectangle, scale=0.5, fill=black] at (constrain0_coord) (constrain0) {};
\node[rectangle, scale=0.5, fill=black] at (dynamics1_coord) (dynamics1) {};
\node[circle, scale=0.5, fill=black] at (soft_u1_coord) (soft_u1) {};
\node[rectangle, scale=0.5, fill=black] at (constrain1_coord) (constrain1) {};
\node[rectangle, scale=0.5, fill=black] at (dynamics2_coord) (dynamics2) {};
\node[circle, scale=0.5, fill=black] at (soft_u2_coord) (soft_u2) {};
\node[rectangle, scale=0.5, fill=black] at (constrain2_coord) (constrain2) {};
\node[rectangle, scale=0.5, fill=black] at (dynamics3_coord) (dynamics3) {};
\node[circle, scale=0.5, fill=black] at (soft_u3_coord) (soft_u3) {};
\node[rectangle, scale=0.5, fill=black] at (constrain3_coord) (constrain3) {};
\node[rectangle, scale=0.5, fill=black] at (dynamics4_coord) (dynamics4) {};
\node[circle, scale=0.5, fill=black] at (soft_u4_coord) (soft_u4) {};
\node[rectangle, scale=0.5, fill=black] at (constrain4_coord) (constrain4) {};
\node[rectangle, scale=0.5, fill=black] at (dynamics5_coord) (dynamics5) {};
\node[circle, scale=0.5, fill=black] at (soft_u5_coord) (soft_u5) {};
\node[rectangle, scale=0.5, fill=black] at (constrain5_coord) (constrain5) {};
\node[rectangle, scale=0.5, fill=black] at (dynamics6_coord) (dynamics6) {};
\node[circle, scale=0.5, fill=black] at (soft_u6_coord) (soft_u6) {};
\node[rectangle, scale=0.5, fill=black] at (constrain6_coord) (constrain6) {};
\node[rectangle, scale=0.5, fill=red] at (cross1_coord) (cross1) {};
\node[rectangle, scale=0.5, fill=red] at (cross2_coord) (cross2) {};
\end{tikzpicture}}
        \caption{Factor graph}
    \end{subfigure}
    \begin{subfigure}{\linewidth}
    \centering
        \resizebox{0.9\linewidth}{!}{\begin{tikzpicture}
\coordinate (x1_coord) at (1.2, 0);
\coordinate (x0_coord) at (0.0, 0);
\coordinate (u0_coord) at (0.6, -1);
\coordinate (x2_coord) at (2.4, 0);
\coordinate (u1_coord) at (1.7999999999999998, -1);
\coordinate (x3_coord) at (3.5999999999999996, 0);
\coordinate (u2_coord) at (3.0, -1);
\coordinate (x4_coord) at (4.8, 0);
\coordinate (u3_coord) at (4.2, -1);
\coordinate (x5_coord) at (6.0, 0);
\coordinate (u4_coord) at (5.3999999999999995, -1);
\coordinate (x6_coord) at (7.199999999999999, 0);
\coordinate (u5_coord) at (6.6, -1);
\coordinate (x7_coord) at (8.4, 0);
\coordinate (u6_coord) at (7.8, -1);
\node[scale=0.8, fill=white][circle, inner sep=2.8pt, draw, very thin] at (x1_coord) (x1) {$x_1$};
\node[scale=0.8, fill=white][circle, inner sep=2.8pt, draw, very thin] at (x0_coord) (x0) {$x_0$};
\node[scale=0.8, fill=white][circle, inner sep=2.8pt, draw, very thin] at (u0_coord) (u0) {$u_0$};
\node[scale=0.8, fill=white][circle, inner sep=2.8pt, draw, very thin] at (x2_coord) (x2) {$x_2$};
\node[scale=0.8, fill=white][circle, inner sep=2.8pt, draw, very thin] at (u1_coord) (u1) {$u_1$};
\node[scale=0.8, fill=white][circle, inner sep=2.8pt, draw, very thin] at (x3_coord) (x3) {$x_3$};
\node[scale=0.8, fill=white][circle, inner sep=2.8pt, draw, very thin] at (u2_coord) (u2) {$u_2$};
\node[scale=0.8, fill=white][circle, inner sep=2.8pt, draw, very thin] at (x4_coord) (x4) {$x_4$};
\node[scale=0.8, fill=white][circle, inner sep=2.8pt, draw, very thin] at (u3_coord) (u3) {$u_3$};
\node[scale=0.8, fill=white][circle, inner sep=2.8pt, draw, very thin] at (x5_coord) (x5) {$x_5$};
\node[scale=0.8, fill=white][circle, inner sep=2.8pt, draw, very thin] at (u4_coord) (u4) {$u_4$};
\node[scale=0.8, fill=white][circle, inner sep=2.8pt, draw, very thin] at (x6_coord) (x6) {$x_6$};
\node[scale=0.8, fill=white][circle, inner sep=2.8pt, draw, very thin] at (u5_coord) (u5) {$u_5$};
\node[scale=0.8, fill=white][circle, inner sep=2.8pt, draw, very thin] at (x7_coord) (x7) {$x_7$};
\node[scale=0.8, fill=white][circle, inner sep=2.8pt, draw, very thin] at (u6_coord) (u6) {$u_6$};
\path[->, >=stealth, draw=black, line width=0.3pt] (x0) -- (x1);
\path[->, >=stealth, draw=black, line width=0.3pt] (u0) -- (x1);
\path[->, >=stealth, draw=black, line width=0.3pt] (x0) -- (u0);
\path[->, >=stealth, draw=black, line width=0.3pt] (x1) -- (x2);
\path[->, >=stealth, draw=black, line width=0.3pt] (u1) -- (x2);
\path[->, >=stealth, draw=black, line width=0.3pt] (x1) -- (u1);
\path[->, >=stealth, draw=black, line width=0.3pt] (x2) -- (x3);
\path[->, >=stealth, draw=black, line width=0.3pt] (u2) -- (x3);
\path[->, >=stealth, draw=black, line width=0.3pt] (x2) -- (u2);
\path[->, >=stealth, draw=black, line width=0.3pt] (x3) -- (x4);
\path[->, >=stealth, draw=black, line width=0.3pt] (u3) -- (x4);
\path[->, >=stealth, draw=black, line width=0.3pt] (x3) -- (u3);
\path[->, >=stealth, draw=black, line width=0.3pt] (x4) -- (x5);
\path[->, >=stealth, draw=black, line width=0.3pt] (u4) -- (x5);
\path[->, >=stealth, draw=black, line width=0.3pt] (x4) -- (u4);
\path[->, >=stealth, draw=black, line width=0.3pt] (x5) -- (x6);
\path[->, >=stealth, draw=black, line width=0.3pt] (u5) -- (x6);
\path[->, >=stealth, draw=black, line width=0.3pt] (x5) -- (u5);
\path[->, >=stealth, draw=black, line width=0.3pt] (x6) -- (x7);
\path[->, >=stealth, draw=black, line width=0.3pt] (u6) -- (x7);
\path[->, >=stealth, draw=black, line width=0.3pt] (x6) -- (u6);

\path[every node/.style={font=\sffamily\small}]
(x0) edge[-latex, draw=red, scale=2,  line width=0.3pt] node [left] {} (u1);
\path[every node/.style={font=\sffamily\small}]
(x0) edge[-latex, draw=red, line width=0.3pt] node [left] {} (u2);
\path[every node/.style={font=\sffamily\small}]
(x3) edge[-latex, draw=red, line width=0.3pt] node [left] {} (u4);
\path[every node/.style={font=\sffamily\small}]
(x3) edge[-latex, draw=red, line width=0.3pt] node [left] {} (u5);




\path[every node/.style={font=\sffamily\small}]
(x0) edge[-latex,bend left, draw=red, scale=2,  line width=0.3pt] node [left] {} (x2);
\path[every node/.style={font=\sffamily\small}]
(x0) edge[-latex,bend left, draw=red, line width=0.3pt] node [left] {} (x3);
\path[every node/.style={font=\sffamily\small}]
(x3) edge[-latex,bend left, draw=red, line width=0.3pt] node [left] {} (x5);
\path[every node/.style={font=\sffamily\small}]
(x3) edge[-latex,bend left, draw=red, line width=0.3pt] node [left] {} (x6);

\end{tikzpicture}}
        \caption{The Bayes Net after Variable Elimination}
        \label{fig:bayes_net}
    \end{subfigure}
    \caption{\footnotesize{Example cross-time-step constraint in a factor graph. The bottom figure is a Bayes net showing variable dependencies after VE.}}
    \label{fig:cross-time-step-factor-graph}
\end{figure}
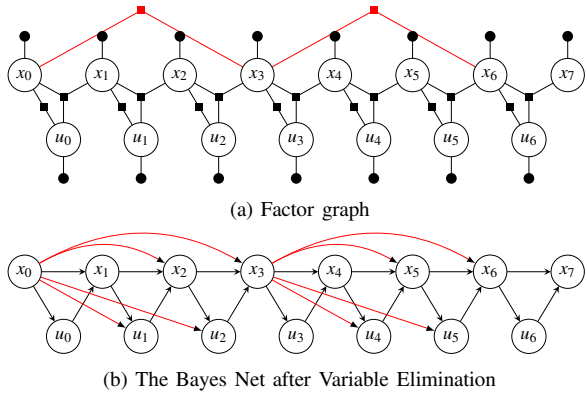
The factor graph's ability to add factors on any set of variables allows us to add more general auxiliary constraints and objectives than \cite{laine2019efficient}, such as cross-time-step constraints. 
Note that cross-time-step \textit{objectives} could also be handled the same way if desired.
The VE algorithm for solving EC-LQR with cross-time-step constraints (or even objectives) remains exactly the same as in Section \ref{sec:ecLQR}. 
For example, in Fig. \ref{fig:cross-time-step-factor-graph}, the cross-time-step constraint is \red{$Sx_{n_c+p} + Sx_{n_c} + s = 0$}. When eliminating $x_{n_c+p}$, its separator will contain $x_{n_c+p-1}$, $u_{n_c+p-1}$ and $x_{n_c}$. After elimination of $x_{n_c+p}$, the new \constrainttogo{} factor will be connected to not only $x_{n_c+p-1}$ and $u_{n_c+p-1}$, but also $x_{n_c}$.  Subsequent elimination steps will generate similar factors.
As a result, after all variables are eliminated, the final feedback controllers for control inputs between $x_{n_c+p}$ and $x_{n_c}$ are functions of two states instead of just the current state. Fig. \ref{fig:bayes_net} illustrates the result in the form of a Bayes Net \cite{dellaert2017factor} where arrows represent the variable dependencies. 


We further show our method maintains linear complexity with the length of the trajectory. Notice in Fig.~\ref{fig:bayes_net} that each cross-time-step constraint spanning from time step $t_a$ to $t_b$ adds additional dependencies of variables $x_k$, $u_k$ ($t_a<k\leq t_b$) on variables associated with the cross-time-step constraint. Therefore, as long as the maximum number of variables associated with a cross-time-step constraint is bounded by $d$, and the maximum number of cross-time-step constraints spanning over any time step is bounded by $q$, the number of variables involved in any elimination step (which contribute to the $\kappa$ constants) is bounded by $3+(d-1)\times q$ thereby bounding the complexity of each elimination operation.

\section{Experiments}\label{sec:experiment}

We run simulation experiments to demonstrate the capability of the proposed method\footnote{Source code is available on \href{https://github.com/paulyang1990/equality-constraint-LQR-compare}{\underline{Github}}}. We implement our method using the Georgia Tech Smoothing And Mapping (GTSAM) toolbox \cite{dellaert2012factor}. We compare our approach with three baseline methods.  Baseline method 1 is \cite{sideris2011riccati}, Baseline method 2 is \cite{laine2019efficient}, and Baseline method 3 is using Matlab's \texttt{quadprog} quadratic programming solver (which does not produce an optimal control \textit{policy}). 
We first present comparison experiments for EC-LQR on random systems. We then show our approach handling cross-time-step constraints on an example system motivated by a single leg hopping robot.
\begin{figure}
\vspace{0.5em}
    \includegraphics[width=\linewidth]{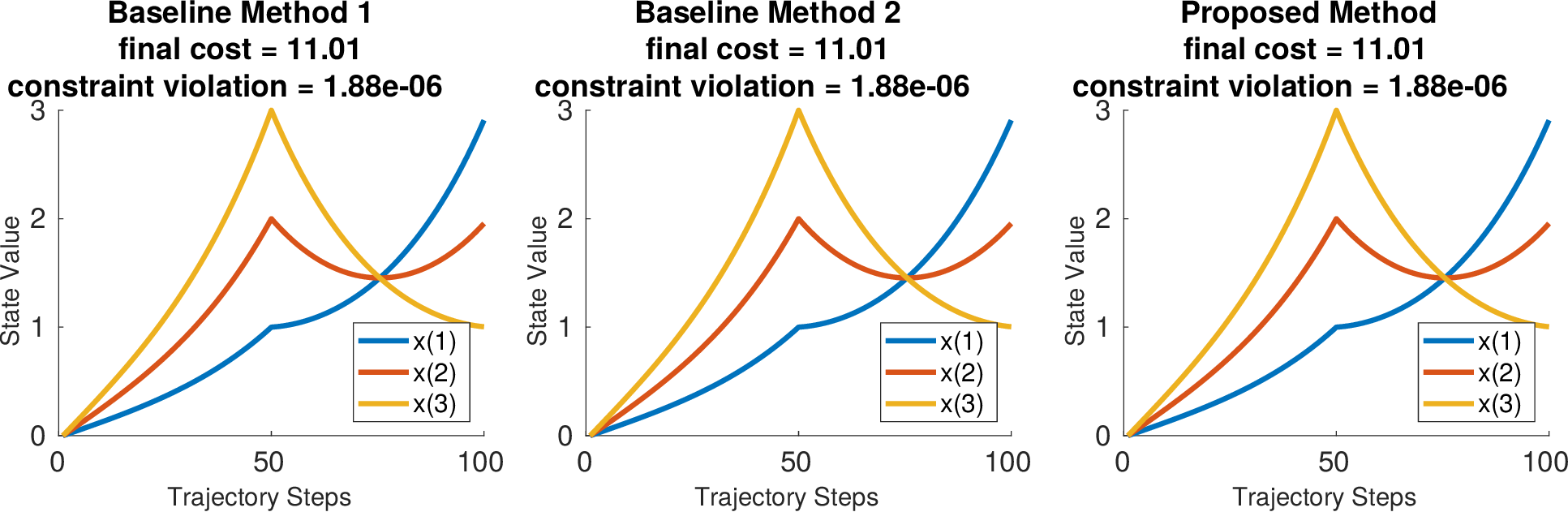}
    \caption{\footnotesize{Optimal trajectory, cost, and constraint violation comparison of three methods for Problem \ref{exp_problem3}. For each method we plot the three dimensions of the state $x$. All methods produce the same result. }}
    \label{fig:compare_3_methods}
\end{figure}
\begin{figure}
    \centering
    \includegraphics[width=\linewidth]{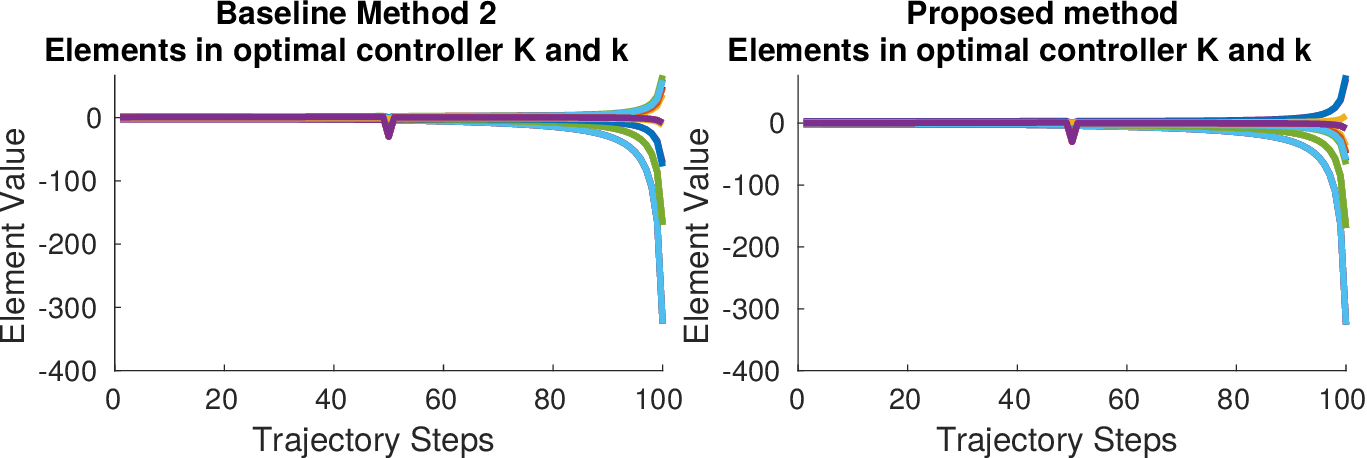}
    \caption{\footnotesize{The plots of feedback control gain matrices from Baseline Method 2 and ours (we omit Baseline 1 because its result is identical to Baseline 2). Each curve represents one element in $K_t$ or $k_t$.}}
    \label{fig:close-loop}
\end{figure}
\subsection{Cost, Constraint Violation \& Controller Comparison}
The first experiment is to find the optimal trajectory for a simple system with $x_i\in\mathbb{R}^3$ and $u_i\in\mathbb{R}^3$ that is subject to state constraints.  The EC-LQR problem is given by:
\begin{subequations}\label{exp_problem3}
\begin{align}
    \min_{\textbf{u}} \ (x_T-&x_N)^TQ_{xx_T}(x_T-x_N) + \sum_{t=0}^{T-1}(x_t^TQ_{xx_t}x_t+u_t^TQ_{uu_t}u_t) \nonumber \\
    \text{s.t. }       \ x_{t+1} & = F_{x} 
                                    x_t 
                                  + F_{u} 
                                    u_t, \quad x_0  = [0\ \ 0\ \ 0]^T, \nonumber\\
                      x_N &= [3\ \ 2\ \ 1]^T, \quad x_{T/2}  = [1 \ \ 2 \ \ 3]^T \label{eqn:state-only-constraint}
\end{align}
\end{subequations}
where $dt = 0.01$, $F_{x}  = I_{3\times3}+I_{3\times3}\cdot dt$,  $F_{u} =I_{3\times3}\cdot dt$, $T = 100$, $Q_{xx_t} = 0.01\cdot I_{3\times3}$, $Q_{uu_t} = 0.001\cdot I_{3\times3}$, and $Q_{xx_T}  = 500\cdot I_{3\times3}$. In this case $\mathcal{C} = \{0,T/2\}$.

Fig. \ref{fig:compare_3_methods} compares the optimal state trajectories using three methods.  Baseline 3 is omitted for space reasons, but all three baselines and our method arrive at the exact same solution, with 0 constraint violation and identical total cost, as expected since the optimal solution is unique.

To show our method can also handle state and control local constraints, we replace the last state-only constraint (\ref{eqn:state-only-constraint}) to be a constraint that contains both the state and the control as $x_{N/2} + u_{N/2} + [1\ \ 2\ \ 3]^T= 0$.
We solve this problem to get the optimal controllers $u_t=-K_tx_t+k_t$. $K_t$ and $k_t$ are identical among Baseline 1, Baseline 2 and ours. Fig. \ref{fig:close-loop} omits Baseline 1 for space reasons.  Baseline 3 does not produce a controller.

\subsection{Run Time Comparison}
We focus on comparing our method and Baseline 2 since Baseline 2 is the only baseline that has linear complexity and generates a feedback policy. Both methods are implemented in C++ and tested on a computer with an Intel i7-8809G 3.10GHz CPU. We generate random problems with given sizes and compare average run times over 10 trials. With $l_t=m-1$ dimensional local constraints at every time step, we first fix $n=m=3$ and vary trajectory length $T$:\\
\begin{tabular}{|c|c|c|c|c|c|c|}
    \hline
     $T$ & 100 & 200 & 300 & 400 & \parbox[t][][t]{\widthof{1.234}}{\centering500} &\parbox[t][][t]{\widthof{1.234}}{\centering600} \\
    \hline
    \cite{laine2019efficient} (ms)  & \textbf{0.88} & \textbf{1.06} & \textbf{1.67} & \textbf{2.01} & \textbf{2.35} & \textbf{2.81}\\
    \hline 
    Ours (ms) & 2.32 & 3.17 & 4.30 &4.68 & 5.86 & 6.86\\
    \hline 
\end{tabular}\vspace{0.2em}\\
then we fix $T=100$ and increase $n$ and $m$ together:\vspace{0.2em}\\
\begin{tabular}{|c|c|c|c|c|c|c|}
    \hline
     $n,m$ & 10 & 20 & 30 & 40 & 50 & 60 \\
    \hline
    \cite{laine2019efficient} (ms)  & \textbf{3.74} & 14.5 & 44.1 & 83.5 & 152.3 & 247.7\\
    \hline 
    Ours (ms) &3.81 & \textbf{11.8} & \textbf{27.1} & \textbf{51.2} & \textbf{99.0} & \textbf{170.2}\\
    \hline 
\end{tabular}\vspace{0.2em}

The experiments show that for both methods, run time grows linearly with increasing trajectory length as expected. 
Our method performs better for larger state and control dimensions.  We believe this behavior is attributable to QR factorization being faster than SVD (used in Baseline 2), which overcomes the graph overhead for large $m$.

\subsection{Cross-time-step Constraints}
\crossnewtrue 

To illustrate an example of how cross-time-step constraints can be used to generate useful trajectories, we use a double integrator system ($x_i = [\mathrm{position}; \mathrm{velocity}], u=\mathrm{acceleration}$) with periodic ``step placements''.  Consider the x-coordinate of a hopping robot's foot which initially starts in contact with the ground and makes contact with the ground again every 20 time steps.  Each contact, it must advance forward by 0.6 units and match the ground velocity (which may be non-zero e.g. on a moving walkway).  The problem is given by:
\ifcrossnew 
\begin{subequations}\label{exp_problem5}
\begin{align}
    \min_{\textbf{u}} & \ x_T^TQ_{xx_T} x_T + \sum_{t=0}^{T-1}(x_t^TQ_{xx_t}x_n+u_t^TQ_{uu_t}u_t) \\
    \text{s.t. }       \ x_{t+1} & = \begin{bmatrix} 1 & dt \\ 0 & 1\end{bmatrix}
                                    x_t 
                                  + \begin{bmatrix} 0 \\ dt\end{bmatrix}
                                    u_t \label{eqn:exp_prob51}, \ \
    x_0  = [0\ \ 0]^T,\\
    x_{n_c+20} &-
    x_{n_c} = \begin{bmatrix}0.6 \ 0\end{bmatrix}^T,\quad n_c \mbox{\small$ = 0, 20, 40, 60, 80$} \label{eqn:cross-time-step-constraint}
\end{align}
\end{subequations}
\else 
\begin{subequations}\label{exp_problem4}
\begin{align}
    \min_{\textbf{u}} & \ x_N^TQ_{xx_T} x_N + \sum_{n=0}^{N-1}(x_n^TQx_n+u_n^TRu_n) \\
    \text{s.t. }       \ x_{n+1} & = \begin{bmatrix} 1 & dt \\ 0 & 1\end{bmatrix}
                                    x_n 
                                  + \begin{bmatrix} 0 \\ dt\end{bmatrix}
                                    u_n \label{eqn:exp_prob41}\\
    x_0 & = [0\ \ 0]^T, \ \ x_N = [3\ \ 2]^T \\
    \begin{bmatrix} 0 & 0 \\ 0 & 1\end{bmatrix}& x_{n_c+10} -
    \begin{bmatrix} 0 & 0 \\ 0 & 1\end{bmatrix}x_{n_c} = 0, \ \ n_c = 20k+1 \label{eqn:cross-time-step-constraint}
\end{align}
\end{subequations}
\fi 
\ifcrossnew
    The cross-time-step constraints (\ref{eqn:cross-time-step-constraint}) enforce that contacts must occur at a fixed position relative to and with the same velocities as the previous contacts $p=20$ time steps prior.
    These create constraint factors between two state variables $p=20$ time steps apart, as in Fig. \ref{fig:cross-time-step-factor-graph} ($p=3$ in Fig. \ref{fig:cross-time-step-factor-graph}). 
\else
    The cross-time-step constraint \ref{eqn:cross-time-step-constraint} requires the velocity of the double integrator be the same every other 10 time steps. It creates constraint factors between two state variables $p=10$ time steps apart, as illustrated in Fig. \ref{fig:cross-time-step-factor-graph}.
\fi

\begin{figure}
\vspace*{0.5em}
    \centering
    \ifcrossnew
        \includegraphics[width=\linewidth, clip]{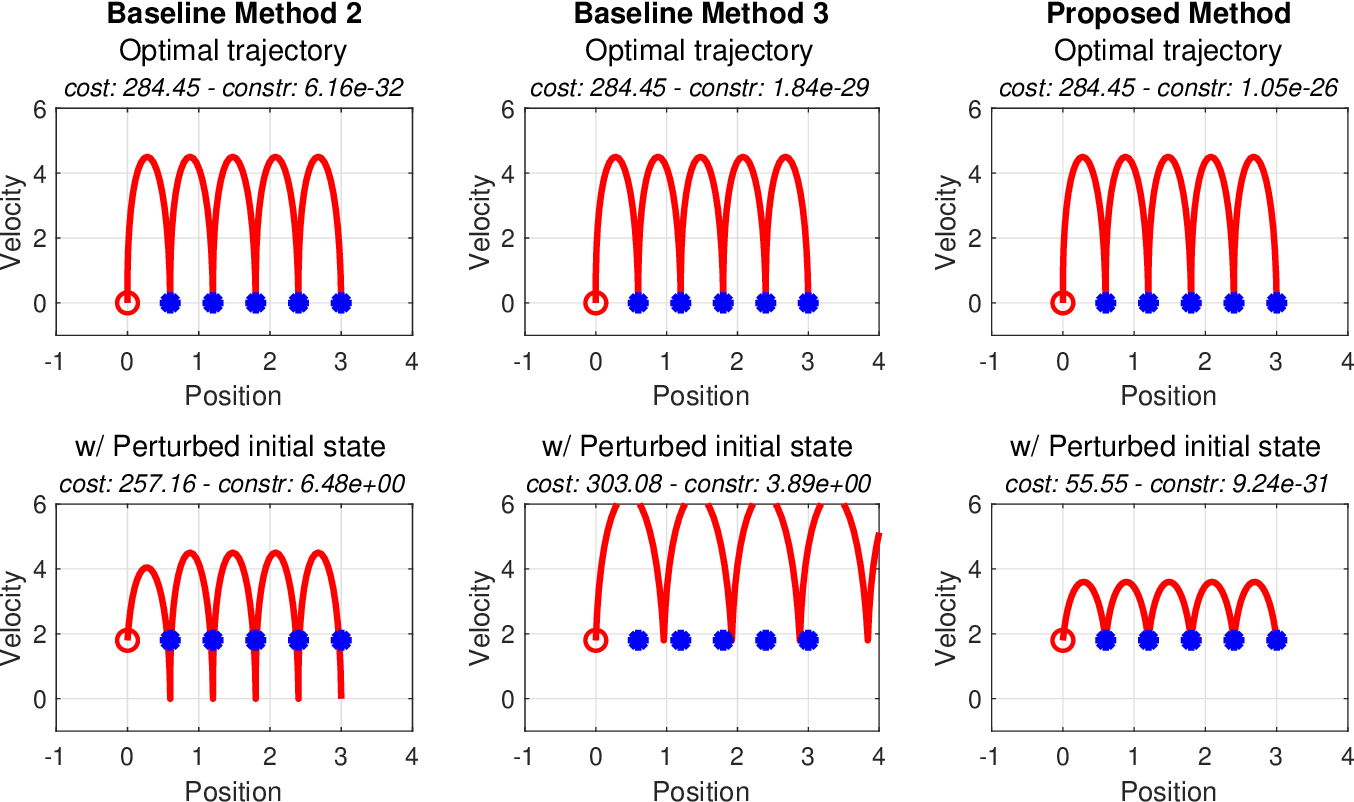}
    \else
        \includegraphics[width=\linewidth]{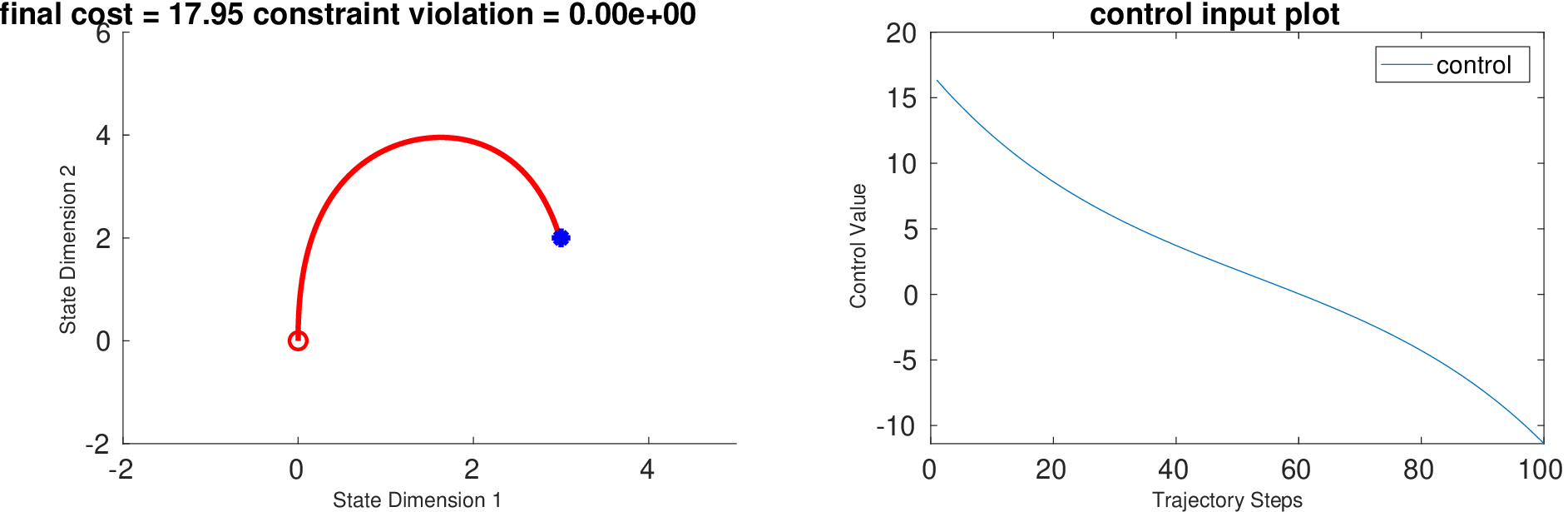}
        \includegraphics[width=\linewidth]{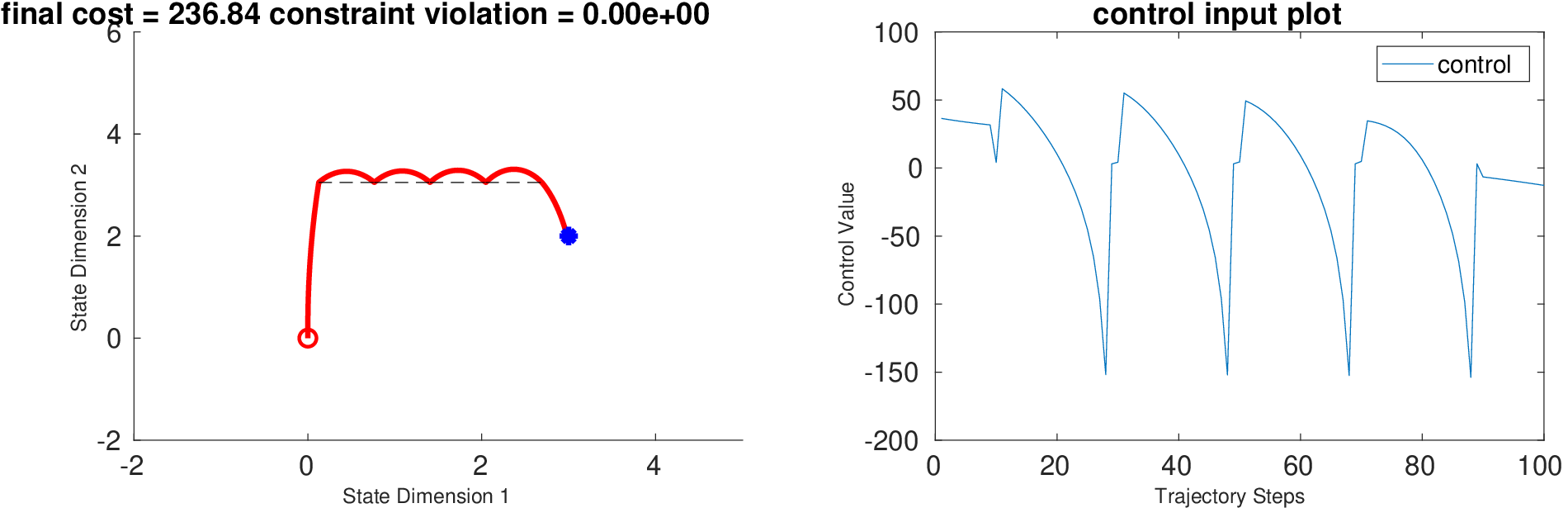}
    \fi
    \caption{\footnotesize{The state trajectories solving Problem \ref{exp_problem5} using Baseline method 2 (left), Baseline method 3 (middle), and our proposed method (right) with control sequence/policies applied to the original problem (top) and after perturbing the initial state (bottom).  All methods generate the same trajectory to the initial problem, but only ours gives a policy which generates the optimal trajectory for the perturbed problem.  
    ``Cost'' and ``Constr'' denote the total objective cost and constraint violation, respectively.
    }}
    \label{fig:cross-time-step-result}
\end{figure}

Fig. \ref{fig:cross-time-step-result} shows the solutions to Problem \ref{exp_problem5} using Baseline 2 \cite{laine2019efficient}, Baseline 3 (QP), and our method, as well as the results when using the same controllers with a perturbed initial state $x_0=[0\ \ 1.8]^T$ (i.e. walking on a moving walkway with velocity 1.8).  We omit Baseline 1 from the Figure for space reasons since it performs identically to Baseline 2.  We apply some modifications to allow for comparison since Baselines 1 and 2 cannot natively handle cross-time-step constraints and Baseline 3 cannot generate an optimal policy, but even so, the adjusted baselines do not generate optimal trajectories from perturbed initial state, as shown in Fig. \ref{fig:cross-time-step-result} (bottom).
For Baseline 2, we convert the cross-time-step constraints to same-time-step constraints $x_{n_c} = [0.03n_c\ 0]^T$ for $n_c=0, 20, \ldots$ resulting in incorrect constraints after perturbing the initial state.  An alternative would be to introduce 10 additional state dimensions (two for each cross-time-step constraint) analagous to Lagrange multipliers, but we argue that such an approach is not sustainable for online operation and many cross-time-step constraints. For Baseline 3, we re-use the control sequence from Problem \ref{exp_problem5} for the perturbed case.  Our method's control law produces a state trajectory that is optimal and without constraint violation even with a perturbed initial state an shown in Fig. \ref{fig:cross-time-step-result} (bottom right).  

\section{Future Work}\label{sec:future_work}

Just as LQR is a building block for Differential Dynamic Programming (DDP) \cite{mayne1973differential, li2004iterative}, linear factor graphs could also be a building block for more general nonlinear optimal control problems. In this direction, the following practical developments should be investigated:
incorporating inequality constraints e.g. using barrier or penalty functions \cite{grandia2019feedback}; 
extending to 
    nonlinear systems using nonlinear factor graphs \cite{dellaert2017factor}; addressing over-constrained ``constraints'' in VE via prioritization of constraints;
    leveraging incremental solving using Bayes Trees \cite{kaess2012isam2} to do efficient replanning;
    and combining estimation and optimal control into the same factor graph  to better close the perception-control loop.


\section{Conclusions}\label{sec:conclusion}
In this paper, we proposed solving equality constrained linear quadratic regular problems using factor graphs.
We showed that factor graphs can represent linear quadratic optimal control problems with auxiliary constraints by capturing the relationships amongst variables in the form of factors.
Variable elimination, an algorithm that exploits matrix sparsity to optimize factor graphs, is used to efficiently solve for the optimal trajectory and feedback control policy. We demonstrated that our approach can handle more general constraints than traditional DP approaches while also matching or exceeding state-of-the-art performance with traditional constraints. We believe our method has great potential to be used in a number of complex robotics systems which require solving more general constrained optimal control problems.


\bibliographystyle{IEEEtran}
\bibliography{sections/references}

\end{document}